\documentclass[review,a4paper,fleqn,authoryear]{cas-dc}
\usepackage[authoryear,longnamesfirst]{natbib}
\usepackage{amsmath,amssymb,amsfonts}
\usepackage{algorithmic}
\usepackage{graphicx}
\usepackage{textcomp}
\usepackage{xcolor}
\newcommand\red[1]{#1}
\usepackage{amsfonts}
\usepackage{bm}
\usepackage{amsmath}
\newtheorem{Def}{Definition}
\usepackage[ruled,norelsize]{algorithm2e}
\usepackage{multirow}
\usepackage{makecell}
\usepackage{subfig}
\usepackage{lineno}
\newcommand\rev[1]{#1}
\newcommand\rjf[1]{}

\newcommand\eg{\emph{e.g.}}
\newcommand\ie{\emph{i.e.}}

\def\tsc#1{\csdef{#1}{\textsc{\lowercase{#1}}\xspace}}
\tsc{WGM}
\tsc{QE}

\begin{document}

\shorttitle{Preference-Agile Multi-Objective Optimization}

\shortauthors{Jin et~al.}
\title[mode = title]{Preference-Agile Multi-Objective Optimization for Real-time Vehicle Dispatching}
\author[1]{Jiahuan Jin}[style=chinese]
\ead{jiahuan.jin@nottingham.edu.cn}
\author[1]{Wenhao Zhao}[style=chinese]
\ead{scywz4@nottingham.edu.cn}
\author[2]{Rong Qu}[style=chinese]
\ead{rong.qu@nottingham.ac.uk}
\author[1]{Jianfeng Ren}[style=chinese]
\ead{jianfeng.ren@nottingham.edu.cn}
\author[1]{Xinan Chen}[style=chinese]
\ead{xinan.chen@nottingham.edu.cn}
\author[3]{Qingfu Zhang}[style=chinese]
\ead{qingfu.zhang@cityu.edu.hk}
\author[1]{Ruibin Bai}[style=chinese]
\cormark[1]
\ead{ruibin.bai@nottingham.edu.cn}

\address[1]{School of Computer Science, University of Nottingham Ningbo China, Ningbo, China.}
\address[2]{School of Computer Science, University of Nottingham, Nottingham, UK.}
\address[3]{Department of Computer Science, City University of Hong Kong, Hong Kong}
\cortext[cor1]{Corresponding author.}

\fntext[1]{
This work was supported in part by the National Natural Science Foundation of China under Grant 72071116, and in part by the Ningbo Municipal Bureau Science and Technology under Grants 2025Z197, 2023Z237.}

\begin{abstract}
Multi-objective optimization (MOO) has been widely studied in literature because of its versatility in human-centered decision making in real-life applications. Recently, demand for dynamic MOO is fast-emerging due to tough market dynamics that require real-time re-adjustments of priorities for different objectives. However, most existing studies focus either on deterministic MOO problems which are not practical, or non-sequential dynamic MOO decision problems that cannot deal with some real-life complexities.  
To address these challenges, a preference-agile multi-objective optimization (PAMOO) is proposed in this paper to permit users to dynamically adjust and interactively assign the preferences on the fly. To achieve this, a novel uniform model within a deep reinforcement learning (DRL) framework is proposed that can take as inputs users' dynamic preference vectors explicitly. Additionally, a calibration function is fitted to ensure high quality alignment between the preference vector inputs and the output DRL decision policy.  
Extensive experiments on challenging real-life vehicle dispatching problems at a container terminal showed that PAMOO obtains superior performance and generalization ability when compared with two most popular MOO methods. Our method presents the first dynamic MOO method for challenging \rev{dynamic sequential MOO decision problems}. 
\end{abstract}

\begin{highlights} 
    \item An end-to-end Preference-Agile Multi-Objective Optimization (PAMOO) is proposed.  
    \item A novel network architecture is tailored with enhanced generalization. 
    \item The proposed method facilitates dynamic and interactive preference adjustment in an online setting. 
    \item The method marks the first Dynamic Multi-Objective Reinforcement Learning for port operations.  
\end{highlights}


\maketitle

\section{Introduction}
Decision-making in complex systems poses significant challenges due to its multi-objectivity, non-linearity, exponential-sized search space, and limited predictability. Despite recent progress in AI, automated decision making underpinned by AI and advanced algorithms remains sporadic in real-life. \rev{The core limitation of existing multi-objective optimization (MOO) methods is that they are designed for either deterministic or non-sequential decision problems where a Pareto optimal set exists and can be pre-computed. Consequently, these methods are unsuitable for sequential decision problems involving real-time decisions based on observations of uncertainties and dynamic user preferences \citep{Tu2025scalarisation}.}

Let's take container ports as an example. 
One of the most commonly overlooked factors is the requirement of dynamic priority adjustment over multiple objectives, representing interests from different stakeholders like shipping companies, port terminal and truck drivers.  
Many existing deterministic models and solution methods for the port terminal optimization \citep{skinner2013optimisation, he2015integrated} are computationally expensive and lack required flexibility and robustness under uncertainties. 
%
In practice, engineers rely on greedy heuristics embedded with experience and domain knowledge \citep{chen2016dynamic}, whose performance often suffers from their myopic nature. Although these heuristics can be enhanced through data-driven hyper-heuristics \citep{zhang2022deep,chen2022cooperative,chen2025deep}, they are designed for single objective optimization. 

\rev{In practice, dynamic MOO is naturally required. During peak times, critical equipment such as quay cranes (QC) must operate at maximum workload capacity. This ensures that vessels spend minimal time at the berth, thereby maximizing the overall throughput. However, during periods of low activity, factors such as labor costs and operational efficiency in the form of truck mileage and energy consumption, become increasingly significant. This implies that decision makers need to dynamically adjust the preferences of different objectives in real-time, and solutions must be provided promptly.}

In addition to classic stochastic programming based methods (e.g., \cite{jiang2021soft}), in recent years, new frameworks have been proposed for optimization under uncertainty. One emerging scheme is the dynamic utilization of risk-aware patterns \cite{xue2025evolutionary,zhang2026online}. Another popular thread is deep reinforcement learning (DRL), which has witnessed flourishing successes for solving real-life combinatorial optimization problems (COP) with uncertainties \cite{bai2023analytics, bengio2021machine, mazyavkina2021reinforcement,tu2023deep}, thanks to the seminal work of the pointer network 
\citep{vinyals2015pointer}, including operation optimization problems at ports, which are then tackled by the DRL methods and its variants \citep{jin2024container}. However, most of these methods are designed for the optimization of a single objective, lacking the versatility to \rev{accommodate dynamic user preferences reflecting the real-time balancing across objectives from different stakeholders}. Meanwhile, the current multi-objective optimization (MOO) methods (e.g., \cite {mosavi2014engineering}) are designed for deterministic problems and lack required flexibility for dynamic preference changes. 

 Technically, this sequential MOO decision problem falls under the category of online multi-objective optimization (MOO) with posterior preferences. To the best of our knowledge, there is currently no existing MOO approach that can be directly used. This is due to two main reasons: (1) Existing preference-based MOO approaches \citep{mosavi2014engineering} are designed solely for deterministic problem settings that disallow uncertainties. (2) While some recent data-driven MOO methods do support a certain level of uncertainties, they either do not support dynamic adjustment of preferences during problem-solving or are too time-consuming for online MOO problems.

\begin{figure*}[htbp]
\centering
\includegraphics[width=1.0\textwidth]{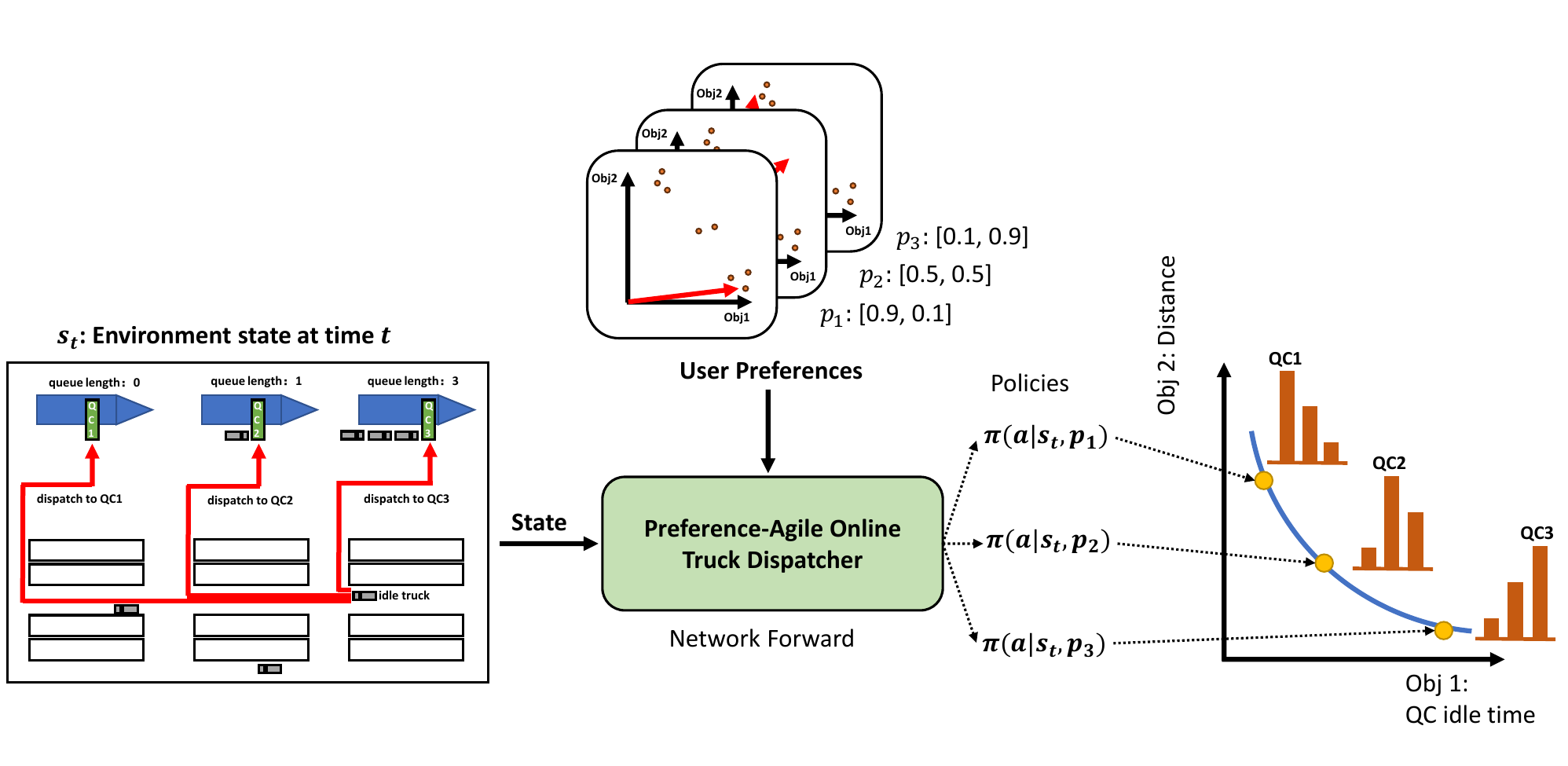}
\caption{\label{fig:pamoo_framework} A simple scenario example to illustrate the proposed \rev{PAMOO algorithm} for online truck dispatching in a container terminal. \rev{At time $t$, one idle truck needs to be dispatched for a new task (dedicated to different QCs). Among three choices QC1, QC2 and QC3 with incremental queue lengths and decremental empty travel distances (indicated by three red lines on the left side of figure). Dispatching decisions are made by jointly considering a dynamic preference (based on expert experience on macro-level situational judgments) and fine-grained data reflecting real-time state features.} 
}
\end{figure*}

To address the aforementioned issues and challenges, we propose a versatile \rev{learning-based algorithm}, namely, Preference-Agile Multi-Objective Optimization (PAMOO) for \rev{dynamic MOO decision making with sequentially revealed uncertainty and dynamic user preference}. The proposed method uses a novel neural network structure within a state-of-the-art reinforcement learning framework. To support the real-time preference adjustments, a dedicated preference embedding is integrated \rev {as a core component} of the network architecture. A non-parameterized calibration function is automatically fitted to ensure good alignments between the generated solutions by PAMOO and the user expectations for given objective preference vectors. A neighborhood-aware attention \rev{mechanism} is used to enhance the discrimination power of different decision-making scenarios and the algorithm's generalization. 
Rather than relying on a finite set of pre-trained dispatching policies, PAMOO adopts a single model and generates decisions with arbitrary preferred trade-offs and permits users to dynamically decide the trade-off in different uncertainty scenarios (see Figure~\ref{fig:pamoo_framework}) and interactively adjust preferences without heavy re-computation (see Figure~\ref{fig:methodology}). 

Our main contributions can be summarized as follows. 
\rev{
\begin{itemize}
    \item We propose a novel network architecture for large-scale online MOO problem (PAMOO). By leveraging the advantages of the advanced multi-objective reinforcement learning (MORL) and unified embeddings of dynamic preferences and state features, our method enables simultaneous learning of various preferences with high sample efficiency, leading to improved diversity and quality of the Pareto solution set. 
    \item The proposed PAMOO algorithm combines a neigh-borhood-aware QC attention module and a preference alignment function to enhance generalization across unseen preference scenarios and problem sizes. 
    \item Our methodology marks the first dynamic Multi-Objective Reinforcement Learning (MORL) integrated with a high-fidelity simulation for real-life truck dispatching for container ports. This breakthrough allows port authorities to dynamically and interactively adjust trade-off decisions, leveraging their expert experience on macro-level situational judgments with fine-grained operational optimization. Our research represents significant progress in enhancing the applicability of optimization methods to complex real-life problems, and it promotes the adoption of more effective and user-centered port management strategies. The source code is available from https://github.com/jjh-port/pamoo.
\end{itemize}
}



\section{Literature Review}\label{sec:review}
\subsection{Multi-Objective Optimization}
\rjf{Reorganize this subsection. What are the main research challenges for MOO? What are the main categories for MOO? How the existing methods address the challenges. Describe existing methods in each category. In the end, you may discuss what are the remaining challenges, possibly our proposed method aims to address.} 

Multi-objective optimization (MOO) has been widely used to model complex optimization problems across different disciplines \citep{Ehrgott2026fifty}. Classical MOO methods like the weighted sum approach, goal programming, and the $\epsilon$-constraint method, often transform multi-objective problems into single-objective ones. However, these techniques often fail to capture the true breadth of the Pareto front. As a result, evolutionary algorithms (EAs) \citep{zhang2007moea}, such as NSGA-II, MOEA/D, and SPEA2, have gained popularity due to their population-based approach and ability to approximate the Pareto front effectively \citep{deb2002fast}. However, these approaches can be computationally prohibitive and are not suitable for real-time decision making under uncertainties. Consequently, pre-trained mechanisms have  been integrated to improve the search efficiency and solution quality \citep{maashi2014multi, zhang2022meta}. Some recent efforts have been made by combining advanced machine learning methods with MOO frameworks \citep{wang2025new,sarkar2025integrating}. 
However, most of these methods have only been applied to canonical problems and their capability in handling real-life problems has not been tested.  


\subsection{Multi-Objective Optimization in Container Terminals} 
Due to the business nature, existing MOO cannot meet the requirements of marine port services perfectly. They are proposed either for deterministic settings \citep{kim2013multi,liu2016bi,hu2019optimal,prayogo2022bi}, or for the multi-objective dynamic control problems \citep{farina2004dynamic,jiang2022evolutionary}, which are not sequential decision problems that can be readily formulated as MDP processes.

\subsection{Multi-Objective Reinforcement Learning}  
Multi-objective reinforcement learning (MORL) \citep{hayes2022practical} is a special topic in RL domain to handle the diversified objectives. MORL can be broadly categorized into two classes: single-policy algorithms and multi-policy algorithms, depending on whether there is only one user preference or many unknown ones. In the single-policy case, the specific preference over objectives is given, and hence the problem is simplified to a single-objective optimization problem. 
Multi-policy approaches aim at generating a set of policies to approximate the Pareto front. These methods can be further divided into outer loop methods and inner loop methods. 
Similar to the decomposition methodology in MOO \citep{zhang2007moea}, outer loop methods derive the policy set by operating on several single-objective problems separately \citep{parisi2014policy}. 
Outer loop methods focus on designing training mechanisms that could accelerate the learning process, \eg, re-use the network parameters of learned models \citep{li2020deep, zhang2022meta} or learn to initialize the network parameters that adapt to different preferences faster \citep{chen2019meta}.  In contrast, inner loop methods are designed to directly produce multiple policies, \eg, multi-objective fitted Q-iteration (MOFQI) \citep{castelletti2011multi}, weight-conditioned network \citep{abels2019dynamic}, and Pareto Q-learning \citep{ruiz2017temporal}. 
Based on this classification, the proposed PAMOO belongs to multi-policy inner loop method. Compared to outer loop methods, the proposed inner loop method adopts a single unified DRL policy network that provides a simpler and yet flexible interface and performs well across different scenarios in container ports. \rjf{Some brief description of the proposed method here.}  

\section{Problem Definition}\label{sec:formulation}
\begin{figure}[htbp]
\centering
\includegraphics[width=0.45\textwidth]{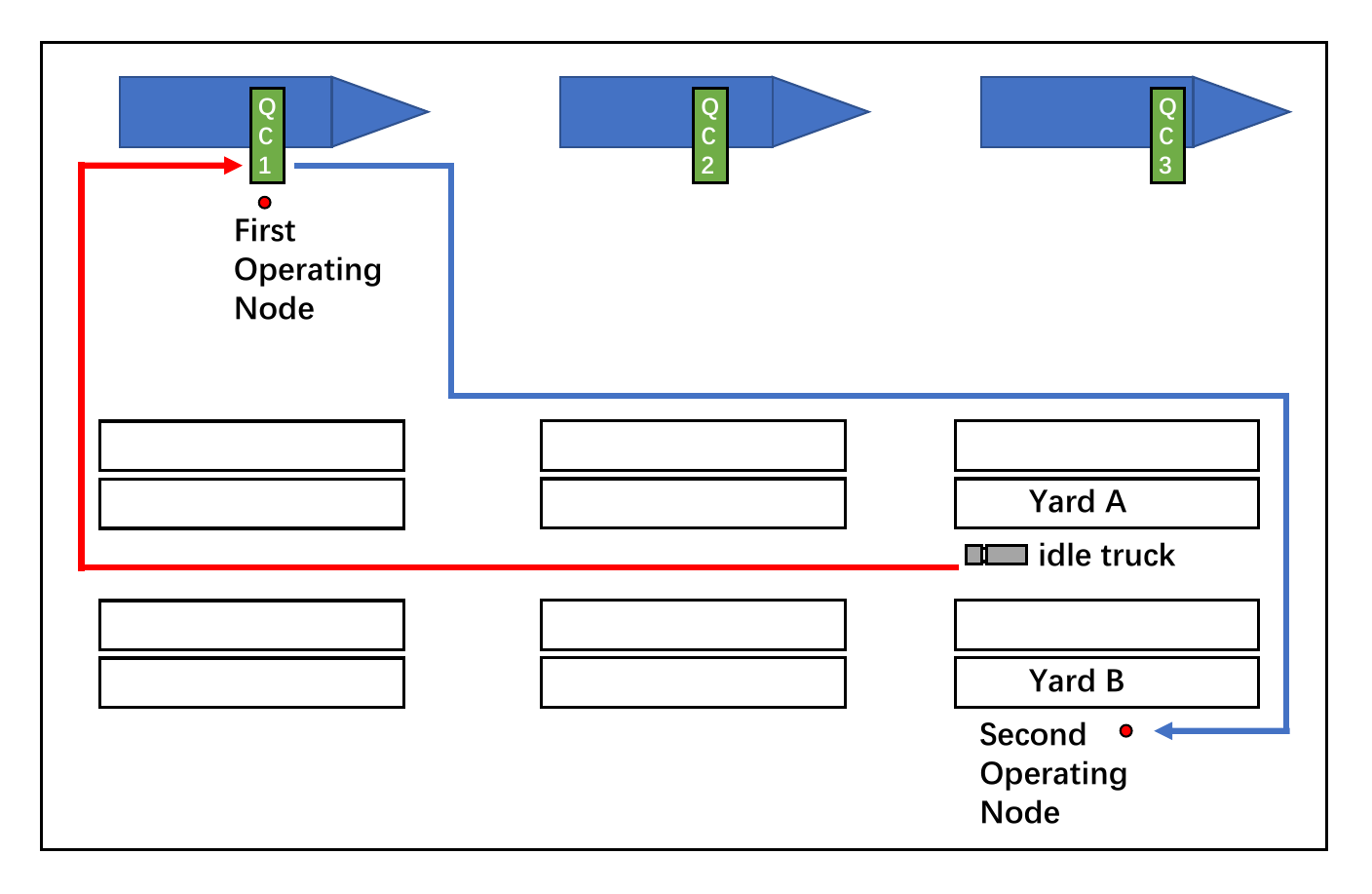}
\caption{\label{fig:task_flow} A route example for a single task. An idle truck receives the dispatching task at yard A. The first and second operation nodes are QC1 and crane at yard B, respectively. The truck route in red represents empty mileage and the route in blue is the loaded travel distance. The objectives are to minimize both aggregated idle time of all QCs and total empty mileages by all trucks. }
\end{figure}

In real-life container terminals, (inner) trucks are used to transport containers between quay cranes (QCs) placed at berth and yard cranes (YCs) at land-side (see Figure~\ref{fig:task_flow}). A standard transportation task can be defined by its first and second operating nodes and its index in the queues. A task can be either from QC to YC (i.e., unloading containers \rev{for importers}) or from YC to QC (loading containers \rev{for exporters}). Upon the arrival of a vessel, the set of containers to be loaded and unloaded can be determined in advance and a fixed number of QCs are assigned to service the vessel \rev{through upstream problems such as berth allocation, QC assignment, and scheduling. The resulting solution to the upstream problems is represented by a set of task lists, each attached to a dedicated QC. In practice, each list contains either import or export containers, but not both. The tasks in each list must be completed in the order defined by that list. While tasks across different QCs are completed in parallel, they are implicitly correlated because they often share the same YCs as one of their operation nodes. When multiple operations arrive at a given YC, a scheduling policy must be adopted to sequence them; in this study, we employ the first-come-first-served (FCFS) heuristic. The entire system is orchestrated by a central system called the terminal operating system (TOS), which has the capability to collect real-time status information on all physical entities (i.e., QCs, YCs, containers, vessels, yard blocks, and trucks) as well as the task lists lining up at each QC. The TOS can also broadcast instructions to all schedulable equipment to complete specific tasks.}

 \rev{Through the TOS system, the truck} dispatch process is repeatedly triggered by a task request from an idle truck. The automated dispatch \rev{agent} (i.e., the algorithm proposed in this paper) takes the current states as inputs (see Section \ref{sec:state}) and evaluates all candidate tasks available at the active QCs, before assigning the most suitable task to this truck which would then visit the first and second nodes (\ie operation points) of the assigned task to complete the transportation. A waiting time is imposed if there are other preceding trucks at any node because both QCs and YCs can only handle unit task each time. Upon completion of the task, the truck becomes idle again and triggers a new request until all tasks are completed. In Figure~\ref{fig:task_flow}, an illustrative example is given, comprising of two legs of the dispatched truck route (red and blue). The truck dispatching requires careful consideration to avoid interruptions of QC operations (QC waiting for the in coming trucks). The examined problem in this study extends several previous single-objective optimization approach, adopting a bi-objective of simultaneously minimizing both the total idle time of QCs and total truck empty travel distance. In order to fully mimic complex business processes and uncertainties, a discrete event-based simulation approach is used to train our proposed PAMOO agent. 
 
The problem is mathematically formulated as follows. Denote $\mathsf{Q}$ and $\mathsf{Y}$ the set of QCs and YCs of the container terminal and $d$ the parking lot of the truck fleet (i.e., depot), $d \notin \mathsf{Q} \cup \mathsf{Y}$. All trucks are initialized in the depot at the beginning of the simulation. Let $\mathsf{N} = \mathsf{Q} \cup \mathsf{Y} \cup d$. The term working instruction refers to a unit-sized transportation task that each truck can maximally handle at any time. The set of all tasks are denoted as $\mathbf{W}$. $\mathsf{W}^q$ indicates the task list of $q^{th}$ QC, $\mathsf{W}^q \in \mathbf{W}$ and $w^q_i$ refers to the $i^{th}$ task in $\mathsf{W}^q$, $w^q_i \in \mathsf{W}^q$. Let $f^q_i$ and $s^q_i$ be the first and second operating location of the task $w^q_i$ and $f^q_i, s^q_i \in \mathsf{Q} \cup \mathsf{Y}$. Denote $\mathsf{V}$ the truck fleet, $|\mathsf{Q}|\leq|\mathsf{V}|\leq|\mathbf{W}|$, and each truck $v$ involves in the the dispatching process. $O^q_i$ and $L^q_i$ denote the QC's operation duration and truck waiting time in the QC queue for task $w^q_i$, $O^q_i > 0, L^q_i \geq 0$, respectively. Denote $D^q_i$ the time step when task $w^q_i$ is dispatched. $T_{init}$ is the start time of the simulation and $T_{end}$ refers to the end time once all the task in $|WI|$ are finished. Several functions that help to model some details during the truck dispatching process are established as below. The function $\tau(x, y)$ and $\delta(x, y)$ are used to query the travel time and distance from location $x$ to $y$ separately where $x, y \in \mathsf{N}$. Expression $\beta (t, v)$ returns the current position of the truck $v$ and $\beta (t, v) = d$ if $t = T_{init}$. The formula $\lambda(w^q_i)$ calculates the yard crane service time (incl. waiting in the yard queue) for a given task $w^q_i$. 

Equation \eqref{dispatch} decides if a specific task $w^q_i$ is assigned to truck $v$. 

\begin{equation} 
\label{dispatch}
\centering
\alpha (w_i^q, v) =
    \begin{cases}
        1, & \text{$w_i^q$ is assigned to $v$} \\
        0, & \text{otherwise}
    \end{cases}
\end{equation}

Expression \eqref{qc_type} indicates the loading or unloading type of a given task $w^q_i$.

\begin{equation}
\label{qc_type}
\centering
\phi (w_i^q) =
    \begin{cases}
        1, & \text{$w_i^q$ is loading task} \\
        0, & \text{otherwise}
    \end{cases}
\end{equation}

Let $T^q_i$ \eqref{arrive_qc} computes the time duration that a truck takes to arrive at the QC in task $w^q_i$.

\begin{equation}
\label{arrive_qc}
\begin{split}
\centering
T^q_i = [\tau(f^q_i,s^q_i) + \lambda(w^q_i)]\phi(w^q_i) + \sum_{v\in \mathsf{V}}\tau(\beta(D^q_i, v), f^q_i)\alpha(w_i^q, v)
\end{split}
\end{equation}

Formally we can have the following problem formulation:
\begin{equation}
\begin{split}
\label{obj_idle_time}
\centering
\min &\sum_{q=1}^{|\mathsf{Q}|}\sum_{i=2}^{|\mathsf{W}^q|} \max(D^q_i + T^q_i - D^q_{i-1} - T^q_{i-1} - L^q_{i-1} -O^q_{i-1},0) \\ 
& +\sum_{q=1}^{|\mathsf{Q}|}T_1^q
\end{split}
\end{equation}
\begin{equation}
\label{obj_distance}
\centering
\min \sum_{q=1}^{|\mathsf{Q}|}\sum_{i=1}^{|\mathsf{W}^q|}\sum_{v\in \mathsf{V}}\delta(\beta(D^q_i, v), f^q_i)\alpha(w_i^q, v)
\end{equation}

\emph{s.t}

\begin{equation}
\label{positive_distance}
\centering
\delta(x, y) > 0 \quad \forall x \neq y, \quad x, y \in \mathsf{N}
\end{equation}

\begin{equation}
\label{positive_time}
\centering
\tau(x, y) > 0 \quad \forall x \neq y, \quad x, y \in \mathsf{N}
\end{equation}


\begin{equation}
\label{dispatch_to_one}
\centering
\sum_{v \in \mathsf{V}} \alpha(w_i^q, v)=1  \quad \forall w_i^q \in \mathbf{W}
\end{equation}

\begin{equation}
\label{dispatch_first}
\centering
D^q_1 = T_{init} \quad \forall q \in [1, |\mathsf{Q}|]
\end{equation}

\begin{equation}
\label{dispatch_in_order}
\centering
D^q_i \geq D^q_{i-1} \quad \forall q \in [1, |\mathsf{Q}|]
\end{equation}

\begin{equation}
\label{operate_in_order}
\centering
D^q_i + T^q_i + L^q_i \geq D^q_{i-1} + T^q_{i-1} + L^q_{i-q} + O^q_{i-1}\quad \forall q \in [1, |\mathsf{Q}|]
\end{equation}

\begin{equation}
\label{positive_queuing_time}
\begin{split}
\centering
& L^q_i = D^q_{i-1} + T^q_{i-1} + L^q_{i-q} + O^q_{i-1} - D^q_i - T^q_i \\
& if \quad D^q_i + T^q_i < D^q_{i-1} + T^q_{i-1} + L^q_{i-q} + O^q_{i-1}\quad
\end{split}
\end{equation}

\begin{equation}
\label{zero_queuing_time}
\begin{split}
\centering
& L^q_i = 0  \quad\ if \quad D^q_i + T^q_i\geq D^q_{i-1} + T^q_{i-1} + L^q_{i-q} + O^q_{i-1}\quad
\end{split}
\end{equation}

\begin{figure}[htbp]
\centering
\subfloat[]{
\includegraphics[width=0.45\textwidth]{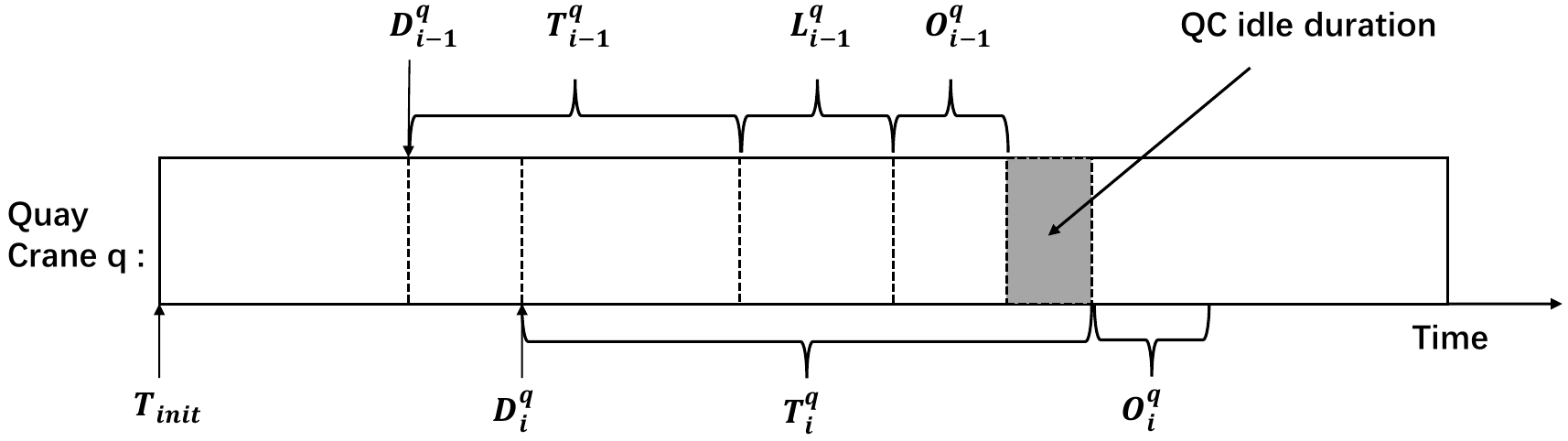} \label{fig:idle_time}
} \\
\subfloat[]{
\includegraphics[width=0.45\textwidth]{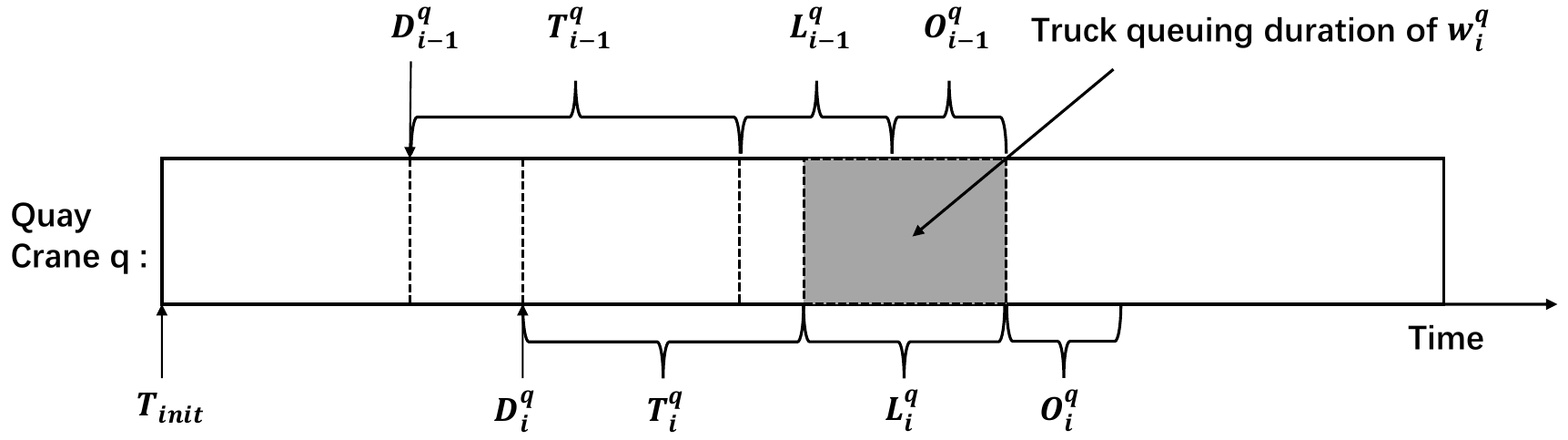} \label{fig:queue_time}
}
\caption{An illustrative example where truck with task $w^q_i$ arrives too late at $q$-th QC, causing a QC idle duration (a) and arrives at QC before the prior task's completion, resulting in truck queuing (b).} 
\end{figure} 


The first objective \eqref{obj_idle_time} minimizes the total QC idle time, hence implicitly maximizes the total throughput. There are some special time steps during a task's completion cycle. The relative positions of these time steps between two adjacent tasks are used to calculate the first objective value (see more details in Figure~\ref{fig:idle_time}). Function \eqref{obj_distance} defines the second objective that aims to minimize the total truck empty travel distance. Constraints \eqref{positive_distance} and \eqref{positive_time} ensure that the traveling distance and time are positive for any two different node in $\mathsf{N}$. Equation \eqref{dispatch_to_one} guarantees each task in $\mathbf{W}$ need to be dispatched and is dispatched to exactly one truck. Equation \eqref{dispatch_first} limits the first task of each QC is dispatched at the beginning of the simulation. Constraints \eqref{dispatch_in_order} makes sure that tasks are dispatched in the pre-defined order. Expression \eqref{operate_in_order} makes sure that each QC must operate tasks in the exactly same order as dispatching and a task starts to be operate by QC only when its prior task are completed. Equation \eqref{positive_queuing_time} computes the time that a truck with task $w^q_i$ needs to wait in the target QC queue when it arrives at the target QC before the completion of the previous task $w^q_{i-1}$. Equation \eqref{zero_queuing_time} ensures that QC operates the task $w^q_i$ immediately if the truck arrives after the completion of the prior task $w^q_{i-1}$ (no queuing duration). 

The intricate constraints \eqref{positive_distance}-\eqref{zero_queuing_time}, coupled with multi-objectivity and uncertainties in travel time and operation times, result in a much more challenging problem that is not sufficiently studied in the research community. Furthermore, one should note that the locations of the nodes in $\mathsf{Q} \cup \mathsf{Y}$ are not stationary because of the movements of both QCs and YCs. Hence the value of $\delta(x,y)$ for any given two nodes varies over time and must be computed in real-time. Due to the stochastic service time of cranes, $O^q_i$ is also subject to uncertainties and is made to follow a probability distribution in our experiments. Finally, the terms $L^q_i$ and $\lambda(w^q_i)$ are also not fully predictable because of the uncertainty of the environment and could only be confirmed after occurrence of the related events. Most of the researchers neglect these details, which limits their practical applications in real-life.

\section{Preference-Agile Multi-Objective Optimization (PAMOO)}\label{sec:methodology}
\begin{figure*}[htbp]
\centering
\includegraphics[width=0.8\textwidth]{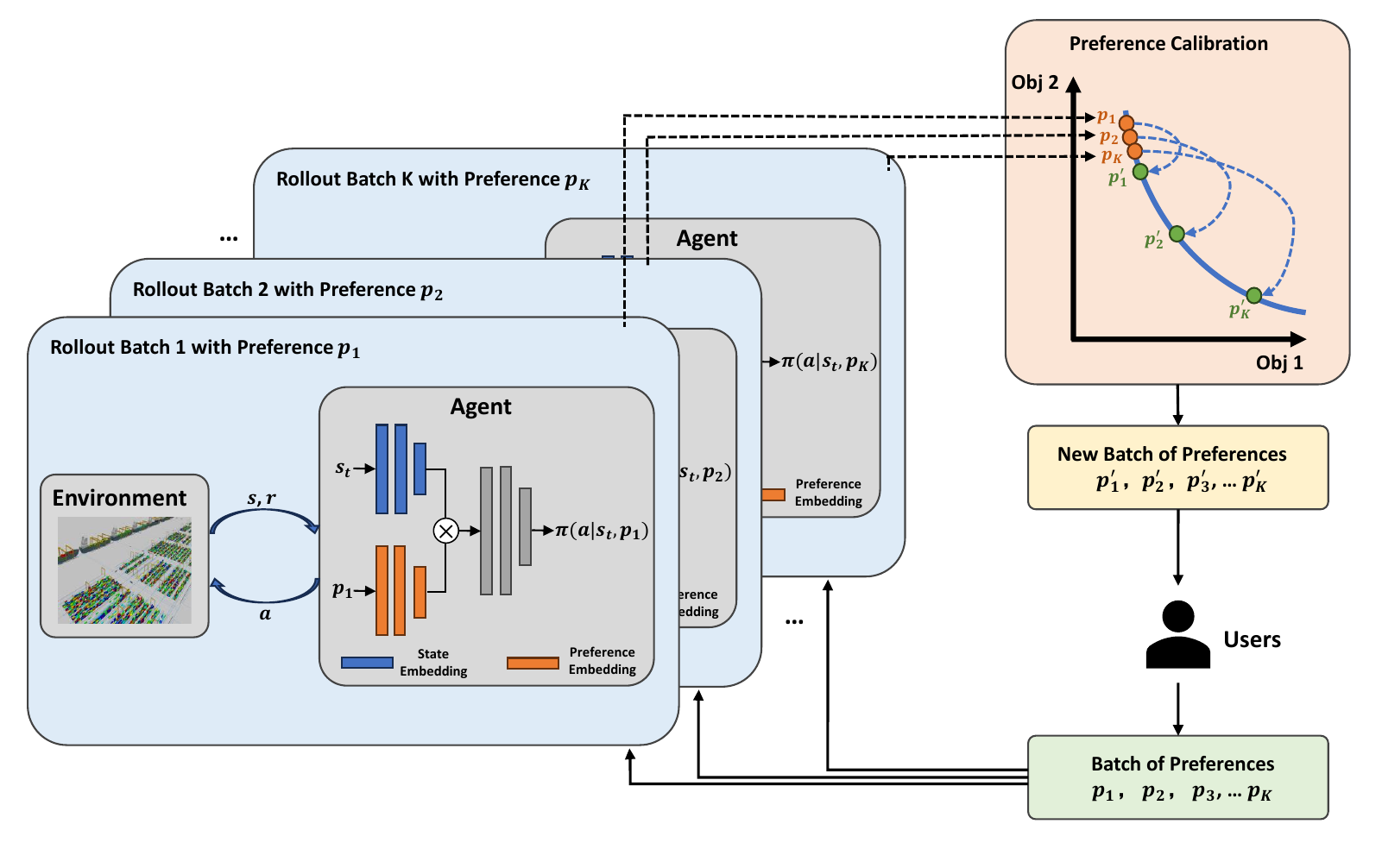}
\caption{\label{fig:methodology} An illustration of learning based interactive adjustments of user preferences in PAMOO. \rev{It employs a multi-policy inner loop RL with policies $\pi^*(a|s,\theta)$. Once trained, PAMOO makes decisions for any combinations of state $s$ and preference vector $\bm{p}$ in a single run. }}
\end{figure*}

In order to effectively handle the complexities and dynamic nature of the concerned real-life MOO problem, our proposed PAMOO method (see Figure~\ref{fig:methodology}) adopts the state-of-the-art multi-policy reinforcement learning framework with fast preference calibration module. We now describe the method in detail.  

\subsection{Formulation of Multi-objective Reinforcement Learning}
\subsubsection{Single-Objective Reinforcement Learning}
Reinforcement learning (RL) is a machine learning paradigm that trains an agent to take intelligent actions through interactions with the environment. The decision making process is often formulated as Markov decision process (MDP). \rev{In a single-objective setting, an MDP is described by the tuple $(S, A, r, \gamma, P)$ where $S$ is the state space and $A$ defines the set of candidate actions. $r$ is a real-valued immediate reward signal associated with the single objective. $\gamma \in [0, 1]$ describes the discount rewards of the MDP and $P$ defines the probabilities of state transitions.} 
%

%
%


\rev{The purpose of the RL agent is to find a policy 
that could map the possible state vectors $s$ to proper actions $a$ in order to maximize the expectation of the accumulated rewards. }

\subsubsection{Multi-Objective Reinforcement Learning}
The multi-objective reinforcement learning (MORL) could be formulated as a
multi-objective Markov decision process (MOMDP) which is denoted as tuple $(S, A, \bm{R}, \gamma, P)$. Compared to the single-objective MDP, the reward function $\bm{R}(s, a, s^{\prime})$ is a vector $\bm{r} \in \mathbb{R}^d$, where $d$ is the number of objectives. Rather than a scalar reward in a single-objective MDP, the vector-valued reward $\bm{r}$ indicates the immediate reward signals for all objectives at the current time step $t$. Therefore, $\bm{V}^{\pi}$ denotes the state-independent value function vector of a given policy $\pi$ in the case of MOMDP and \rev{$V^{\pi}_i$ defines the $i^{th}$ objective, as follows:
\begin{equation}
\centering
V^{\pi}_i = E_s(V^{\pi}_i(s))= E(G^t_i | S_t = s)
\end{equation}
where $G^i_t =\sum^T_{t^{\prime}=1}\gamma R_{t+t^{\prime}}$,  $T$ is the total time steps. When $\gamma$ approaches 1, the agent treats immediate rewards and the possible rewards in the future equally, while prioritizes myopic decisions if $\gamma$ is close to 0. Commonly in policy-based RL method, the policy $\pi(a|s,\theta)$ is defined as a deep neural network with trainable parameters $\theta$. } 

To evaluate a policy $\pi$ in MOMDP, the concept Pareto Dominance \citep{hayes2022practical} is introduced and defined as follows. 

\begin{Def}[Pareto Dominance]
A policy $\pi$ is said to Pareto dominate another policy $\pi^{\prime}$
if and only if $\pi$'s value vector is at least as high $\pi^{\prime}$'s in all objectives and is strictly higher in at least one objective, i.e.,
 $ \pi \succ \pi^{\prime} \Leftrightarrow \forall i, V^{\pi}_i \geq V^{\pi^{\prime}}_i \wedge \exists i, V^{\pi}_i > V^{\pi^{\prime}}_i $
\end{Def}

In the field of multi-objective optimization, the examined objectives are usually conflicted, therefore, MORL aims to learn a set of Pareto optimal policies rather than to obtain a policy that generates optimal solution with regard to one single objective.

A common approach of MORL is to convert multi-objectives to a single objective by using a scalarization function $g(\bm{p}, \bm{o})$, where $\bm{p}$ is the given preference vector among different objectives which satisfies $\sum_{i=1}^{|\bm{p}|} p_i = 1$ and $\bm{o}$ is the corresponding objective vector. In this study, the Weighted Tchebycheff aggregation is used instead of the popular Weighted-Sum scalarization function. The Weighted Tchebycheff aggregation
is defined as follows:

\begin{equation}
\label{eq:tch}
\centering
\red{g_{wt}(\bm{p}, \bm{o}) = \mathop{max}\limits_{1 \leq i \leq |\bm{p}|}\{ p_i|o_i-z_i^*|\},}
\end{equation}
where $z_i^*$ is the ideal value that satisfies $z_i^* < min(o_i)$. The Tchebycheff aggregation tends to have better ability in finding any Pareto optimal policies when provided with adequate number of preference weights.

As mentioned above, our methodology belongs to a multi-policy inner loop RL, which aims to obtain a set of Pareto optimal policies $\pi^*(a|s,\theta)$ for any given preference weight vector $\bm{p}$ in a single run. To achieve this, the preference weight is considered as a special part of the state $s$ and a uniform policy is defined as $\pi(a|s, \bm{p}, \theta)$. A well-trained agent must make the most suitable action based on any combinations of $s$ and $\bm{p}$ that minimize the scalarization objective $g_{wt}(\bm{p}, \bm{o})$.

\subsubsection{Truck Dispatching Optimization as MOMDP}
\noindent\textbf{Environment}
A simulation based on AnyLogic software is developed to depict the environment of our industrial collaborator, Meishan Port Terminal. Numerous details and uncertainty factors in real container terminal are implemented to reproduce high-fidelity real-world optimization cases. For example, the truck or equipment speed may be affected by its status (loaded or empty) or the drivers' operation which should not be considered as constants in simulation. Similarly, the service time of a crane to handle a container depends on the moving distance between crane and truck which is also a dynamic factor. What's more, when operations at nearby yards become busier, the traffic congestion may occur, leading to increased travel time. These properties make the static solution plans almost infeasible to execute and highlight the necessity to treat truck dispatching as an online problem. To further demonstrate the feasibility and applicability of our methodology, the implemented simulation tries to reproduce the complete process of the truck dispatching and the related events that may happen during this procedure. The relevant logic design and parameter settings follow our on-the-spot investigation and the advice from our collaborators in Ningbo-Zhoushan port Meishan terminal in order to make the simulation as realistic as possible.

\noindent\textbf{State}\label{sec:state}
State is the observed information at some time steps for the agent to refer and make the decision. Specifically, in this study, decisions are triggered when some truck become idle and need to be assigned a new task until all tasks are assigned. The state contains both spatial and temporal related information and is organized into a feature vector as the inputs of the RL policy network. The list of state items is given in Table \ref{tab:features}. 

\begin{table}[htbp] 
\caption{Items of agent's observation at each time step.}
\begin{center}
\begin{tabular}{|c|l|}
\hline
Index & Description \\
\hline
1 & The remain task number of each QC \\
\hline
2 & The total working truck number of each QC \\
\hline
3 & The distance for the truck to complete each tasks \\ 
\hline
4 & The necessary distance before the truck arrive at \\ &  each QC \\
\hline
5 & The queue lengths at each QC \\
\hline
6 & The queue lengths at each yards in the \\ & corresponding tasks \\ 
\hline
7 & The transport task type: loading or  unloading \\ &operation \\
\hline
8 & The total heading truck number to each QC \\
\hline
9 & The distance to the each target yards \\ 
\hline
10 & Unique five-bit one-hot encoding for each QC. \\ 
\hline
\end{tabular}
\label{tab:features}
\end{center}
\end{table}

\noindent\textbf{Action}
The actions are problem-specific. In our experiments, it is defined as the specific transport task that is assigned to the current dispatchable truck. As mentioned above, all the tasks associated with each QC are organized in a pre-defined order, which means only the first unassigned task in the list is available for assignment at any time step. Therefore, the algorithm only needs to assign a QC for the requesting truck, therefore, the size of the action space is equal to the total number of non-empty QCs. 

\noindent\textbf{Reward}
This work aims at simultaneously minimizing total QC idle time and total truck empty travel mileage. When the action is determined, the assigned task $w_i^q$ is confirmed. Therefore, the agent receives two types of rewards: the reward related to QC idle time is set to be $-T_1^q$ if the dispatched task is $w_1^q$ and $-\max(D^q_i + T^q_i - D^q_{i-1} - T^q_{i-1} - L^q_{i-1} -O^q_{i-1},0)$ for remaining tasks ($i>1$). The reward about truck empty travel distance is $-\delta(\beta(D^q_i, v), f^q_i)$. Both rewards are the feedback for the action of dispatching the idle truck $v$ to task $w_i^q$. Since this is the minimization problem, negative signs are added to discourage high objective values in both terms. Note that the empty travel distance $-\delta(\beta(D^q_i, v), f^q_i)$ could be computed immediately after implementing the action while the resulting QC idle time $-\max(D^q_i + T^q_i - D^q_{i-1} - T^q_{i-1} - L^q_{i-1} -O^q_{i-1},0)$ is not an immediate reward in an episode. Therefore, the reward for QC idle time is calculated retrospectively at the end of the episode.

\noindent\textbf{State Transition}
A state transition occurs when a truck just becomes idle and triggers a dispatch request. As described in \ref{sec:state}, the state at a time step $t$ is the related information between the target truck and all QCs. Therefore, the state transitions is a sequence of spatial switching among different idle trucks. Moreover, the transitions are also affected by the uncertainties in the environment such as non-deterministic crane service time and truck traveling speed. 


\subsection{Network Architecture}
\begin{figure}[htbp]
\centering
\includegraphics[width=0.5 \textwidth]{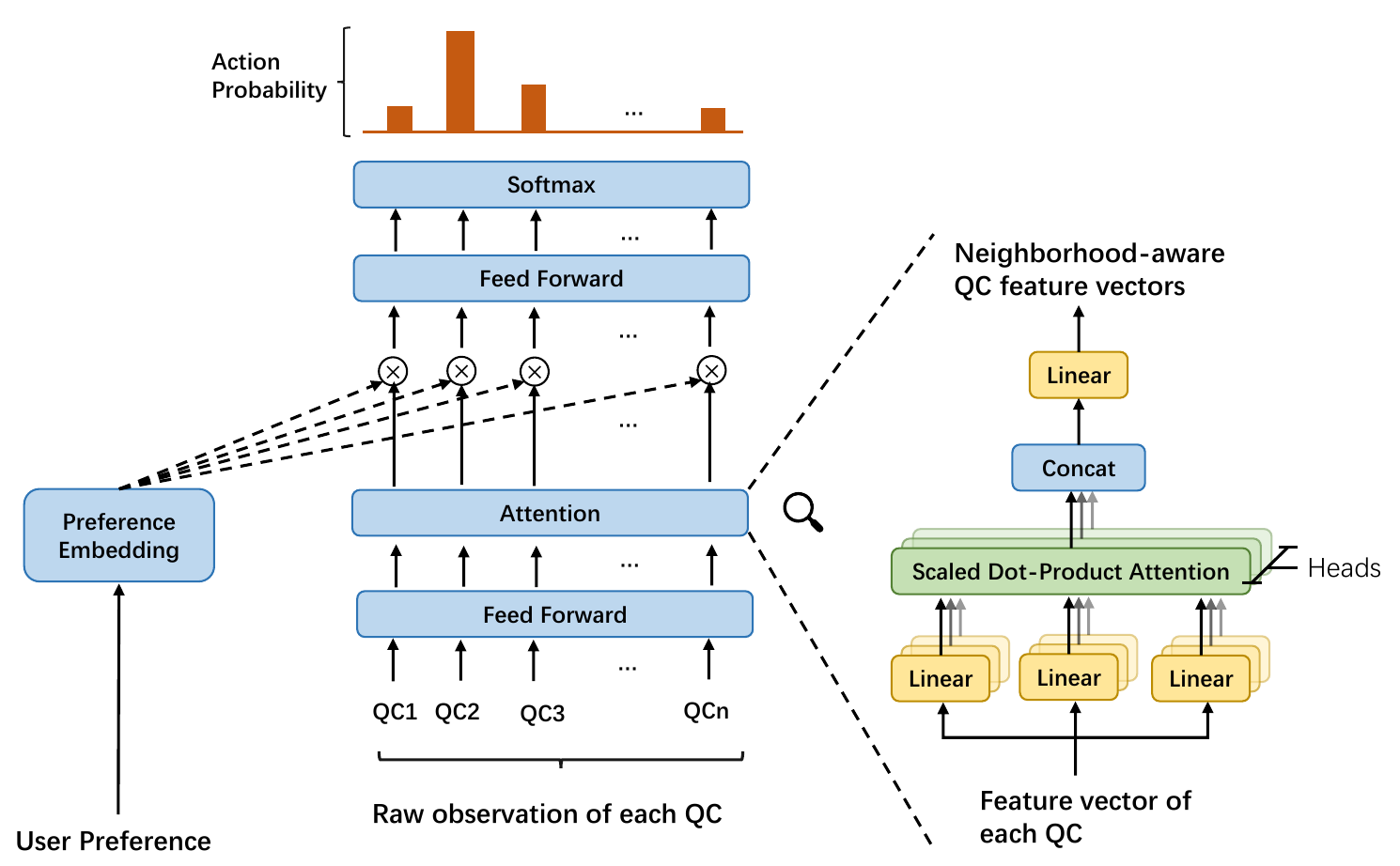}
\caption{\label{fig:network} The network structure of the \rev{proposed PAMOO} for online truck dispatching.}
\end{figure}

Similar to the most of the RL approaches, the truck dispatch policy $\pi(a|s,\theta)$ is approximated by a customized deep neural network with parameters $\theta$. Structure of the policy network used in this research is depicted in Figure~\ref{fig:network}. It can be seen that the proposed network takes both the raw observation of the state features described above and the user preference weight vector as the inputs and outputs the action probability distribution across all candidate actions available for this given preference.  

Firstly, raw observations are fed into a feed-forward layer to generate a 128-dimensional feature vector for each QC. Then a multi-head attention block (following the same structure in \citep{vaswani2017attention}) is adopted to generate neighborhood-aware QC feature vectors, each of which is then combined with the preference vector, generated from the preference embedding layer (a two-layer fully-connected block) by taking the preference weights as the input. After this process, the QC feature vectors that incorporate preference information go through a feed forward layer to map each QC vector to a scalar and together with a softmax layer to generate a probability distribution. 

Unlike most other multi-objective reinforcement learning  studies, our method does not consider user defined preference as a homogeneous feature component in agent's observation but instead treats it separately in the network. The preference information is merged into the QC vectors through Hadamard product operations. The reason for doing so is to make sure the preference weights play a more prominent role such that the dispatching policies are sufficiently sensitive to the changes of user preference vectors. If the preference weights are concatenated to the raw state observations, based on our initial experiments, the network tends to overlook this part in the training or is insensitive to the changes of preferences, which must be prevented. In contrast, \rev{Hadamard product creates a multiplicative feature interaction that allows the important feature to act as a conditioning signal or feature-wise modulator on intermediate representations.} Furthermore, such operation could also selectively highlighted some part in a QC feature vector since the agent may focus on different features when faced with various preference weights.

Apart from the capability to handle multi-objective preference weights, the network treats the inputs as a dynamic set of QC feature vectors. When a QC's task list become empty, the feature vector of this QC needs to be eliminated from the input so that the corresponding QC is excluded from the candidate actions. The multi-head attention block is used to leverage this thanks to its abilities to properly handle the variable input length and its great perception performance of different part of the features. 

\subsection{Learning Strategies for PAMOO}\label{algorithm}
The well-known proximal policy optimization (PPO) \citep{schulman2017proximal} is adopted to train the dispatching agent. PPO has demonstrated its superior performance on convergence and sample efficiency due to the mechanism of reusing the sampling data. 

Denote $s_0^k,a_0^k,\bm{r}_1^k ,...,s_{H-1}^k,a_{H-1}^k,\bm{r}_H^k$ a trajectory of the agent's exploration in the $k^{th}$ episode in training and $K$ is the total number of episodes used in training. $\bm{r}_t^k$ indicates the reward vector for all objectives at time step $t$ of episode $k$ and $H$ is the total steps in an episode which equals to the total number of tasks of the episode in this study. The total scalar reward with preference weight $\bm{p}$ is defined by the following Weighted-Tchebycheff scalarized aggregation. 

\begin{equation}
\label{total_reward}
\centering
R^k = \sum_{t=1}^{H}{g_{wt}(\bm{p},\bm{o})^t} 
\end{equation}
where $g_{wt}(\bm{p},\bm{o})^t$ is the weighted-Tchebycheff function (see eqn. \eqref{eq:tch}) at time step $t$.  
To enhance the stability of the training process, a shared baseline is used to compute an advantage function value for each $s_t^k, a_t^k$ pair as follows: 

\begin{equation}
\label{advantage_function}
\centering
A(s_t^k, a_t^k)= R^k - \frac{1}{K}\sum_{k=1}^{K}R^k
\end{equation}

The probability ratio is defined by Eq. \eqref{theta_ratio}, where the $\theta_{old}$ are the policy network parameters before updating. 

\begin{equation}
\label{theta_ratio}
\centering
Ratio_t^k(\theta) = \frac{\pi(a_t^k|s_t^k, \theta)}{\pi(a_t^k|s_t^k, \theta_{old})}
\end{equation}

PPO optimizes a surrogate objective with clipped probability ratio. A clipped advantage function is defined by Eq. \eqref{clipped_advantage}, where $\epsilon$ is a hyper-parameter ($0 < \epsilon < 1$).

\begin{equation}
\label{clipped_advantage}
\begin{split}
\centering
A^{clip}(s_t^k, a_t^k) = & \min[Ratio_t^k(\theta) A(s_t^k,a_t^k), \\ 
& clip(Ratio_t^k(\theta),1-\epsilon,1+\epsilon) A(s_t^k,a_t^k)]
\end{split}
\end{equation}

Each update of the policy network parameters uses a sampled mini-batch data ($s_t^k, a_t^k$) pairs and their advantage function values with batch size $B$. Each update optimizes the policy with same preference weight $\bm{p}$. The estimated gradient of PPO loss for a particular preference weight is indicated by Eq. \eqref{loss}.

\begin{equation}
\label{loss}
\begin{split}
\centering
\nabla\mathcal{L} = \frac{1}{KH}\sum_{k=1}^K\sum_{t=1}^H A^{clip}(s_t^k,a_t^k)\nabla\log P_{\theta}(a_t^k|s_t^k, \bm{p})  
\end{split}
\end{equation}

\begin{algorithm}[htbp]
    \caption{PPO for preference-agile multi-objective optimization}
    \label{alg:PPO}
    \KwIn {number of iterations $K$, collect $N$ episodes per iteration, steps per episode $T$, $M$ epochs per iteration, clipping rate $\epsilon$, batch size $B (B < NT)$, preference set $\mathcal{P}$}
    Initialize: a differentiable truck dispatch policy parameterization $\pi(a|s, \bm{p}, \theta)$\;
    
    \For{i=1 : K}{
        Randomly select a preference weight $\bm{p}$ from $\mathcal{P}$\;
        \For{n=1 : N}{
            Collect an episode $s_0,a_0,r_1 ,...,s_{T-1},a_{T-1},r_T$, following $\pi(\cdot|\cdot,\bm{p}, \theta)$\;
            Compute each $r_t(a_{t-1}, s_{t-1})$ based on the reward design, $\forall t \in [1, T]$\;
            Compute $R^n$ for preference $\bm{p}$ based on Equation (\ref{total_reward})\;
        }
        Compute $A(s_t^n, a_t^n)$ through Equation \eqref{advantage_function}, $\forall t \in [0, T-1], \forall n \in [1, N]$\;
        Compute PPO loss $\mathcal{L}$ with clipping rate $\epsilon$, optimize the parameters $\theta$ with $M$ epochs and batch size $B$
               
    }
\end{algorithm}

\begin{figure}[htbp]
\centering
\includegraphics[width=0.25 \textwidth]{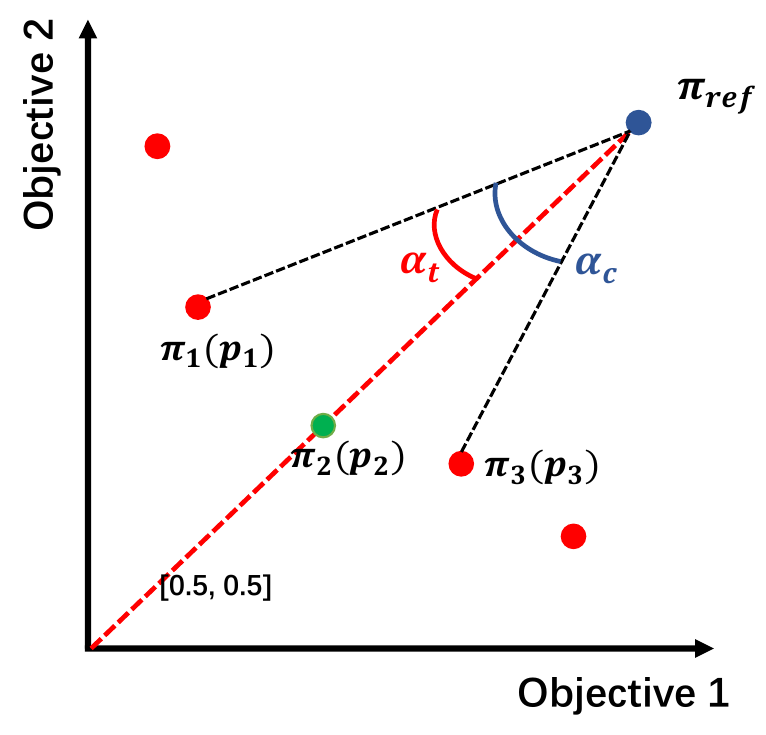}
\caption{\label{fig:pref_adjust} Interpretation of the preference calibration method.}
\end{figure} 

Benefiting from the generalization ability of deep neural networks, a well-trained PAMOO agent is able to make proper dispatching decisions under arbitrary (including unseen) preferences. Intuitively, a set of evenly distributed preference weights would be used to generate an even and dense Pareto front. However, for numerous real-world applications, evenly distributed preferences always fail to generate a regular approximate Pareto front. To tackle such issue, we proposed a simple heuristic approach to calibrate the preference weights based on the relative locations of a set of policies in the objective space. The idea of the preference calibration method is illustrated in Figure~\ref{fig:pref_adjust}, where a set of non-dominated policies (indicated by red points) generated by evenly distributed preference are given. Suppose a policy with preference [0.5, 0.5] which should ideally be located at the direction vector of [0.5, 0.5] (indicated by red dashed line) but the actual location drifts off to some extents. Therefore, this preference needs to be adjusted to make the new policy (a possible policy is $\pi_2$ with preference $p_2$ which is indicated by green points) is located in the correct direction. Firstly, two policies $\pi_1$ and $\pi_3$ with preferences $\bm{p_1}$ and $\bm{p_3}$ which are closest to the target direction [0.5, 0.5] are selected. The area between $\pi_1(\bm{p_1})$ and $\pi_3(\bm{p_3})$ is considered as the neighborhood of the target direction [0.5, 0.5]. The ratio of the angles $\alpha_t$ and $\alpha_c$ are computed. Then, the calibrated preference $\bm{p_2}$ could be obtained by Eq. \eqref{eq:pref_adjust}.

\begin{equation}
\label{eq:pref_adjust}
\centering
\bm{p_2} = \bm{p_1} + \frac{\alpha_t}{\alpha_c}(\bm{p_3} - \bm{p_1})
\end{equation}

This calibration method estimates the preference adjustment amount based on the changes of the angles of two policies locates at the neighborhood area of a particular direction. The implementation details are given in Section \ref{sec:preference-calib}.

\subsection{Uncertain handling}
\rev{Traditionally, stochastic programming (SP) and robust optimization (RO) are popular approaches for handling uncertainty. In SP, problems are often formulated as a two-stage or multi-stage decision process. The early stages are concerned with strategic or tactical decisions that are either impossible or too costly to reverse while late stages correspond to operational decisions. Uncertainties are discretized into a finite set of “scenarios”. SP focuses on optimizing the expectation of strategic and/or tactical objectives by taking feedback from the operational level. The main challenge of SP is the enormous size of the search space that leads to intractable computation, and its lack of efficient mechanisms to leverage real-time data \citep{bai2014stochastic}. Whereas SP optimizes the expectation across all scenarios, RO aims to seek performance guarantees in the worst-case scenario. Solutions by RO are often considered overly conservative and are adopted primarily in safety-critical domains.

In this work, uncertainties are jointly represented via (1) a simulation-based environment with pre-defined distributions of uncertainty variables; (2) vectorized embedding for dynamic user preference weights, and (3) a large set of training and testing instances with diverse dynamics and characteristics. This joint uncertainty representation provides the proposed PAMOO with the versatility and performance required to effectively handle real-world complexities and dynamics. }

\rev{Both PAMOO in this paper and SP and RO in literature assume a stationary environment where the distributions of the random variables do not change over time. An emerging research direction involves considering uncertainty with non-stationary distributions. By combining the classic branch-and-price method with machine learning, \cite{zhang2026online} proposed a three-stage “predict-plan-construct” framework which can not only handle non-stationary distributions, but also combine the strength of the stochastic programming in stage-wise decision making and generative policy-based methods (e.g., RL) in utilizing the real-time gathered state information. Extending this work to non-stationary environments represents a promising future direction. }


\subsection{Comparisons of PAMOO with other MOO methods}
In this section, we discuss the key differences between the proposed PAMOO and other MOO methods. 

\textbf{PAMOO vs. EDMOO}: existing EDMOO \citep{farina2004dynamic,jiang2022evolutionary} is mostly designed for non-sequential MOO problems. Although objectives and/or constraints change over time, at any time instant, the problem is formulated as a static MOO and a Pareto optimal set exists. However, 
    our method is proposed for sequential decision problems with uncertainty and dynamic preferences. The goal of PAMOO is to train general solution policies (instead of solution itself) to sequentially build  solutions by automatically adjusting real-time decisions based on the observations of uncertainties and user preferences. Therefore, PAMOO is a generative method, while most existing EDMOO methods are search based methods for direct solutions. 
    PAMOO complements the current EDMOO methods by focusing on sequential MOO decision problems.

\textbf{PAMOO vs. GP based MOO}: PAMOO aims to address sequential decision problems with uncertainties in the form of MDPs. Generally, both reinforcement learning and genetic programming are both considered to be the most suitable methods because their apparent compatibility with MDPs. Our PAMOO extends a policy-based DRL (i.e., PPO) with a customized preference network that achieves superior performance than both NSGA-II and MOEA/D (see Section \ref{sec:experiment}.B), which are extended by replacing search based EA methods with GP to make them MDP compatible. Compared to tree-based policies from GP, DRL demonstrates better perception ability of problem dynamics and can more efficiently handle high-dimensional features and uncover their complex spatial-temporal relationships.   

\textbf{Computational costs}: the superior performance from PAMOO comes with the some computational costs to train the complex neural network. That said, training a tree-based MOO policy using GP is equally computational intensive (albeit with better interpretability), especially for a large training set. Once PAMOO is trained, however, the inference time under different preferences and uncertainties can be negligible, making itself ideal for practical decision problems that require near real-time responses. 




\section{Experiment}\label{sec:experiment}

\subsection{Experiment Setup}
\subsubsection{Instances}

Each of the instances used in this study consists of 28 QCs, 100 yards, and 700 tasks. The number of trucks varies between different instances and the impact of this variation shall be analyzed shortly. Yard blocks for import and export containers are fixed. In order to enhance the degree of mutual exclusivity of the two objectives, the import and export yards are both evenly distributed among the entire container terminal based on some initial investigation. Due to the dynamic nature and the high-level uncertainties in the environment, the same dispatching policy cannot guarantee same results in terms of two objectives but fluctuates within a small range. Therefore, the $i^{th}$ objective of a policy $\bm{V}^{\pi}$ is evaluated by the average objectives over $C$ sampling, which is shown in Eq. \eqref{n_average}.

\begin{equation}
\label{n_average}
\centering
V_i^{\pi} = \frac{1}{C}\sum_{n=1}^{C}\sum_{t=1}^{H} r_{t, (i)}^n 
\end{equation}
$C$ is set to 128 in this work. That is, in testing stage, 128 problem instances (instances generated with fix random seeds which could ensure the reproducible result for the same dispatching policy) are used for evaluation.

\subsubsection{Parameter Settings}
\rev{Since the performance of PAMOO can be sensitive to the choice of hyperparameters, some level of parameter tuning is required. To avoid excessive trial and error, we began with the same parameters used for the single-objective version of the problem \citep{jin2024container} and subsequently fine-tuned them on a small set of instances before settling on the settings given in Table \ref{tab:hyper-params}.}

\begin{table}
    \caption{Hyper-parameters used in PAMOO.}
    \begin{tabular}{|c|c|}
    \hline
         Parameters & Values  \\
        \hline
         batch size $B$  & 32 \\
         \hline
         clipping rate $\epsilon$ & 0.2 \\
         \hline
         discount rate $\gamma$  & 1 \\
         \hline
         epochs per iteration $M$ & 10 \\
         \hline
         number of iteration $K$ & 5000 \\
         \hline
         episodes per iteration $N$ & 10 \\
         \hline
    \end{tabular}
    
    \label{tab:hyper-params}
\end{table}

\subsubsection{Evaluation Metrics}

\begin{figure}[htbp]
\centering
\includegraphics[width=0.25 \textwidth]{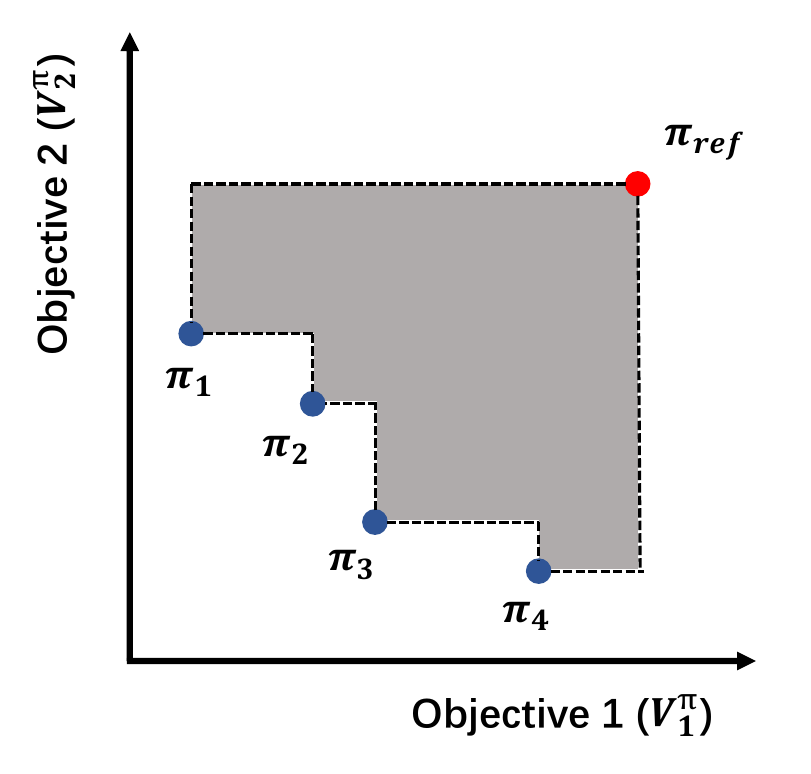}
\caption{\label{fig:HV} The relationship between the hypervolume indicator and the reference policy.}
\end{figure}

The hypervolume (HV) is selected as the performance indicator to measure each method. A reference policy whose value $\bm{V}^{ref}$ is dominated by all other non-dominated policies is firstly needed. For bi-objective optimization, HV measures the space area covered by the non-dominated policies and reference policy over the policy value space. Figure~\ref{fig:HV} illustrates the HV indicator, where the $\pi_1, \pi_2, \pi_3, \pi_4$ indicate the non-dominated policies, $\pi_{ref}$ is the reference policy, and the size of grey area is the value of HV. HV is a common indicator to evaluate multi-objective optimization methods. Generally, a higher HV value implies a better non-dominated policy set as a whole, which indicates the better algorithm performance.

Apart from the HV, the sparsity is also adopted for the evaluation. The sparsity of a given policy set $\mathcal{S}$ with $m$ objectives is defined by Eq. \eqref{sparsity}, where $\widetilde{\mathcal{S}}_j(i)$ is the $i^{th}$ objective value in a partial ordering of these policies sorted by the $j^{th}$ objective. Sparsity evaluates the degree of uniformity of a policy set. A lower sparsity value indicates a more even distribution of policies spread among the entire approximate Pareto front. When evaluating, the HV is considered as the primary focus and when values of HV are comparable, the policy set with lower sparsity is preferred.

\begin{equation}
\label{sparsity}
\centering
Sparsity(\mathcal{S}) = \frac{1}{|\mathcal{S}| - 1}\sum_{j=1}^{m}\sum_{i=1}^{|\mathcal{S}| - 1}(\widetilde{\mathcal{S}}_j(i) - \widetilde{\mathcal{S}}_j(i+1))^2
\end{equation}

Due to the difference in the measurement units by different objectives, the value scale of an objective may vary a lot more than the other, which would cause bias for evaluating Pareto front. To eliminate such effect, both objective values are scaled to range of [0, 1] by using min-max normalization which is defined by Eq. \eqref{min_max_norm}, where $x$ is some objective value of a policy and $X$ is the set of objective values generated by all methods to be compared. When computing HV, the value of reference policy $\bm{V}^{ref}$ is naturally selected as [1, 1]. Therefore, the HV became a value in range of [0, 1].

\begin{equation}
\label{min_max_norm}
\centering
z = \frac{x - min(X)}{max(X) - min(X)}
\end{equation}

\subsubsection{Benchmarks}
To demonstrate the performance of the proposed methodology, two evolutionary algorithms, namely, multi-objective genetic programming (GP) based on paradigms of NSGA-II \citep{deb2002fast} and MOEA/D \citep{zhang2007moea} are implemented, respectively. Traditionally, both NSGA-II and MOEA/D solve the deterministic problems whose solutions can be encoded as a fixed length chromosome vectors (routes, plans, ...etc.) and the operators of genetic algorithm could be adopted during the evolution. Since our problem is an online one, the solutions in our work must be a dispatching rule, with which a solution can be generated. Therefore, both benchmark algorithms are implemented by ourselves. In our proposed method, such a rule is the actor network while in the GP benchmark algorithms, we make it an arithmetic tree \citep{chen2020data} to rank different candidate actions. For both algorithms, the population size is set to 1000. The maximum generation is set to 500, 1000, 2000 respectively in the respective experiments. The probabilities for both crossover and mutation operators are settled at 50\% after some initial trials. For the multi-objective related mechanisms, we followed the original works of NSGA-II \citep{deb2002fast} and MOEA/D \citep{zhang2007moea}. 

\subsection{Comparing with Benchmark Methods}

Table \ref{tab:benchmark} summarizes the comparison results between the proposed PAMOO method with two main stream MOO methods. The performance of HV and sparsity (united by $10^{-4}$) are presented. The gap is measured against the algorithm with highest HV value. For evolutionary algorithms, results after 500, 1000 and 2000 generations of evolutions are reported. For our method, results of 11, 51 and 101 number of evenly distributed preferences are selected as the corresponding comparison. According to the numerical results, our method with 101 preset preference vectors outperforms the other two methods on both HV and sparsity. Overall, the best results (2000 generations) of two evolutionary algorithms have an average gap of 9.74\% towards our method (with 101 preferences) in terms of HV. The performance of NSGA-II and MOEA/D are comparative with each other in general. The HV of MOEA/D is slightly higher than NSGA-II while the sparsity of the NSGA-II is better than MOEA/D. Our method with much less preferences (11 and 51) have average gaps of 0.88\% and 0.41\% respectively, which demonstrates the excellent ability by our method in that even small number of preset preferences can approximate Pareto front very well. 

\begin{table*}[htbp]
\centering
\caption{\label{tab:benchmark}Comparison Result to NSGA-II and MOEA/D.}
\begin{tabular}{|c | c c c | c c c | c c c|}
\hline
\multirowcell{2}{Method} & \multicolumn{3}{c|}{80 Trucks} & \multicolumn{3}{c|}{100 Trucks} & \multicolumn{3}{c|}{120 Trucks} \\
  & HV & Sparsity & Gap(\%) & HV & Sparsity & Gap(\%) & HV & Sparsity & Gap(\%) \\
\hline
NSGA-II (500) & 0.669 & 91.86 & 25.0\% & 0.674 & 31.68 & 23.5\% & 0.708 & 21.73 & 20.45\% \\

NSGA-II (1000) & 0.729 & 19.44 & 18.27\% & 0.728 & 20.1 & 17.37\% & 0.759 & 16.1 & 14.72\% \\

NSGA-II (2000) & 0.807 & 4.01 & 9.53\% & 0.784 & 23.8 & 11.01\% & 0.798 & 9.31 &  10.34\%\\
\hline
MOEA/D (500) & 0.682 & 31.94 & 23.54\% & 0.675 & 75.3 & 23.38\% & 0.693 & 19.75 & 22.13\% \\

MOEA/D (1000) & 0.72 & 35.1 & 19.28\% & 0.746 & 56.85 & 15.32\% & 0.751 & 53.66 & 15.62\% \\

MOEA/D (2000) & 0.819 & 27.85 & 8.18\% & 0.798 & 35.58 & 9.42\% & 0.801 & 27.71 & 10.0\% \\
\hline
Ours (11 pref.) & 0.88 & 42.38 & 1.35\% & 0.857 & 41.88 & 2.72\% & 0.876 & 53.13 & 1.57\% \\

Ours (51 pref.) & 0.89 & 3.67 & 0.22\% & 0.873 & 4.21 & 0.91\% & 0.889 & 4.22 & 0.11\% \\

Ours (101 pref.) & 0.892 & 2.97 & 0.0\% & 0.881 & 2.58 & 0.0\% & 0.89 & 1.67 & 0.0\% \\
\hline
\end{tabular}
\end{table*}

\begin{figure*}[thbp!]
    \centering
    \begin{minipage}[t]{1.0\linewidth}
    \centering
        \begin{tabular}{@{\extracolsep{\fill}}c@{}c@{}c@{}@{\extracolsep{\fill}}}
            \includegraphics[width=0.3 \linewidth]{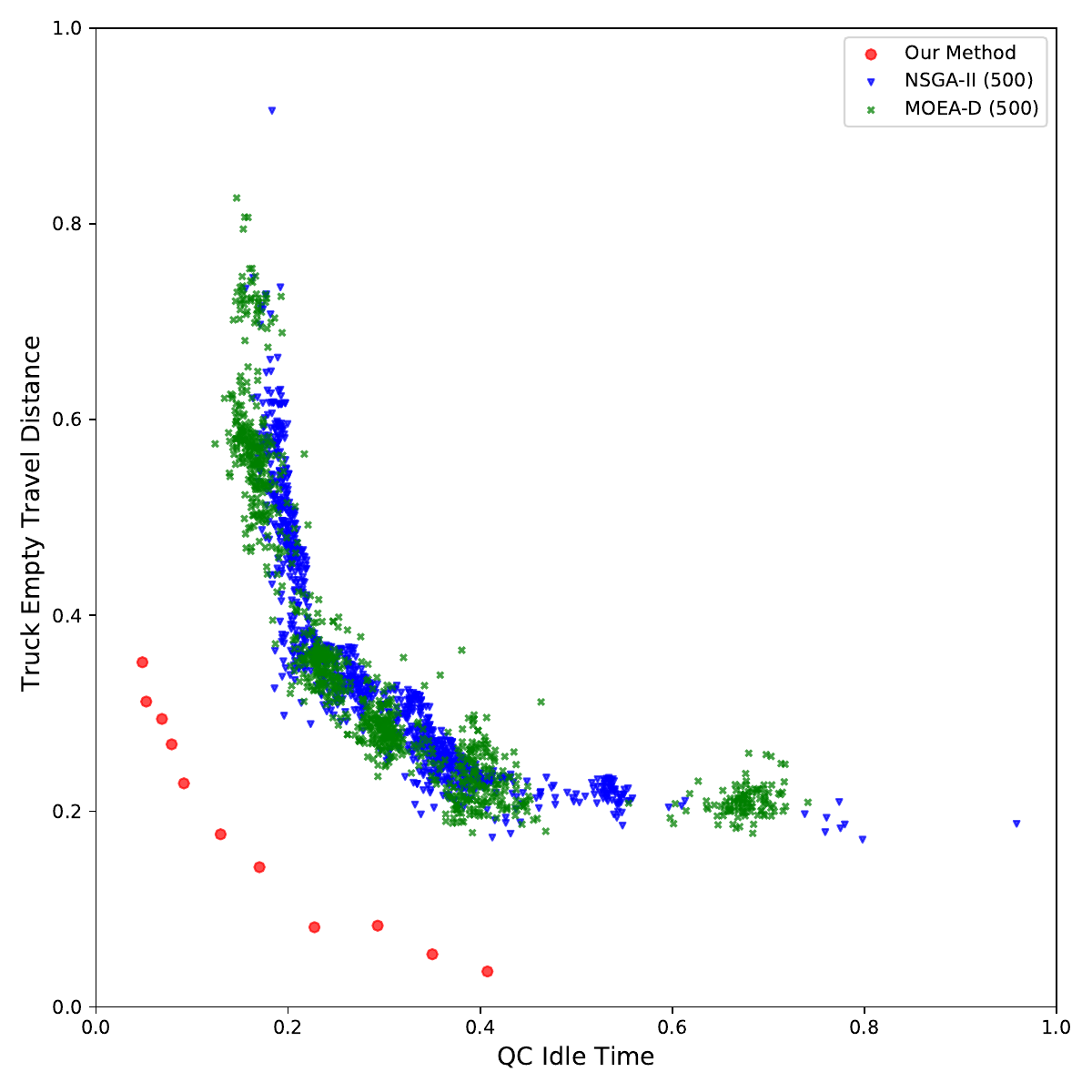} &
            \includegraphics[width=0.3 \linewidth]{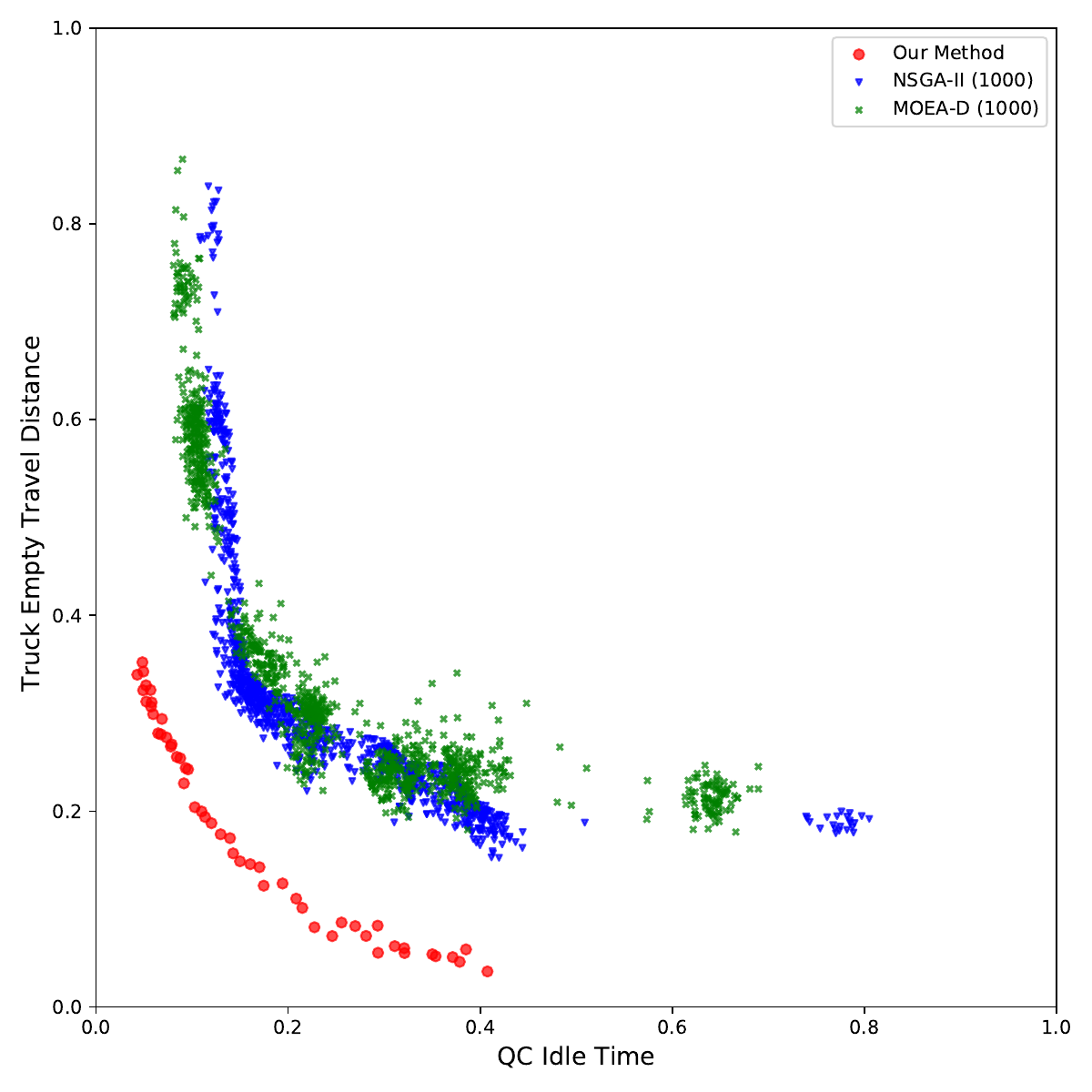} &
            \includegraphics[width=0.3 \linewidth]{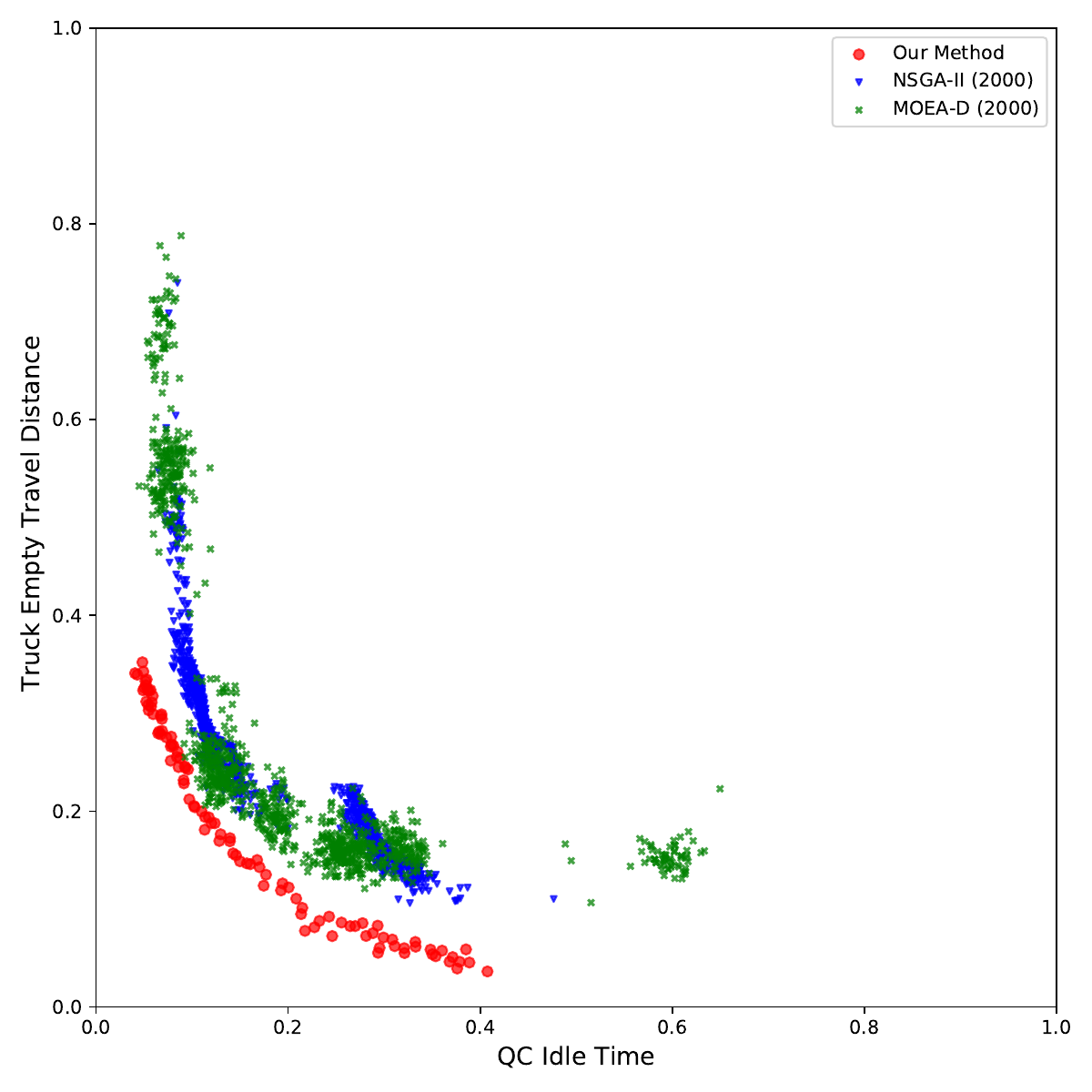} \\
            (a) 80 Trucks, 500 Generations & (b) 80 Trucks, 1000 Generations  & (c) 80 Trucks, 2000 Generations \\ \\
        \end{tabular}
    \end{minipage}
    \begin{minipage}[t]{1.0\linewidth}
    \centering
        \begin{tabular}{@{\extracolsep{\fill}}c@{}c@{}c@{}@{\extracolsep{\fill}}}
            \includegraphics[width=0.3 \linewidth]{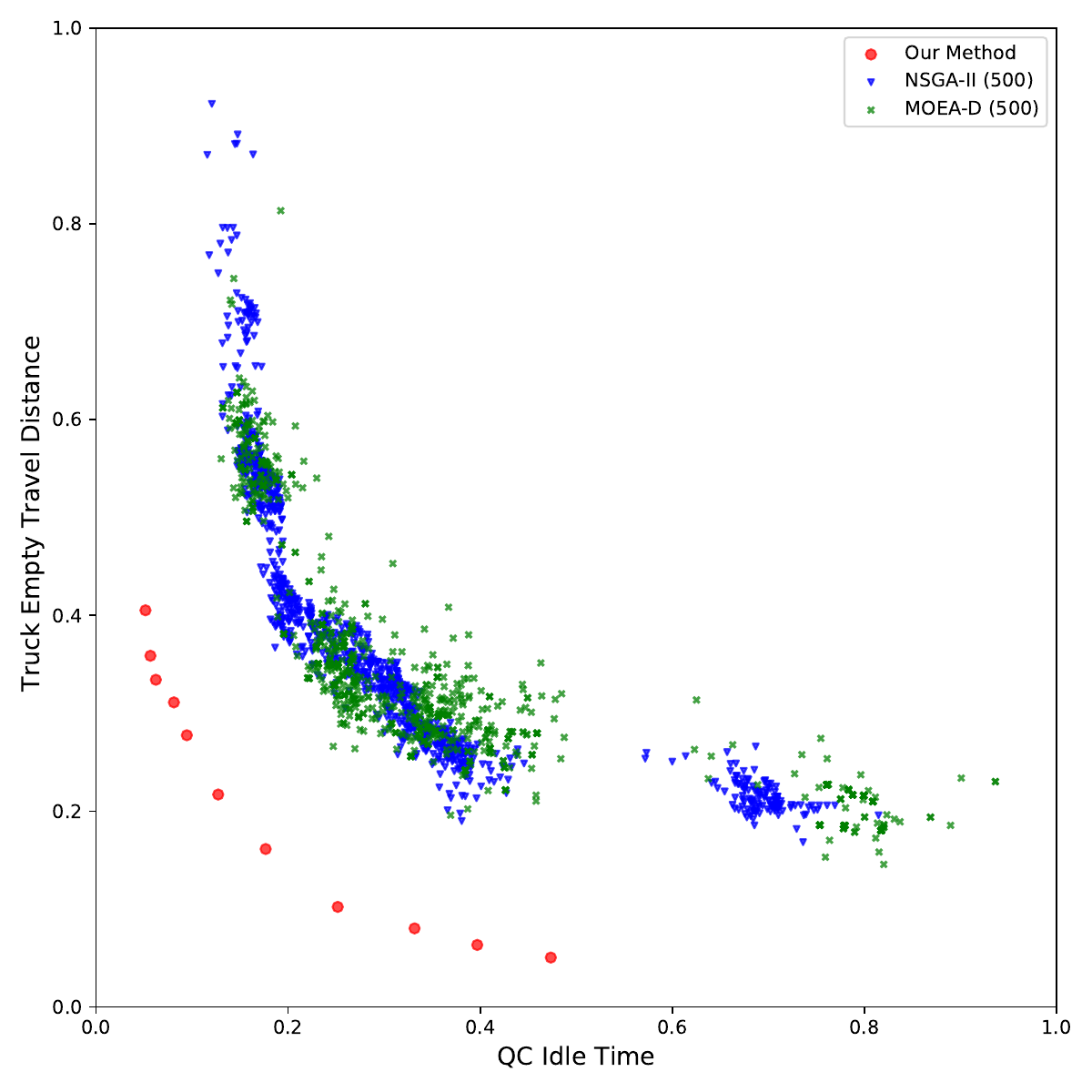} &
            \includegraphics[width=0.3 \linewidth]{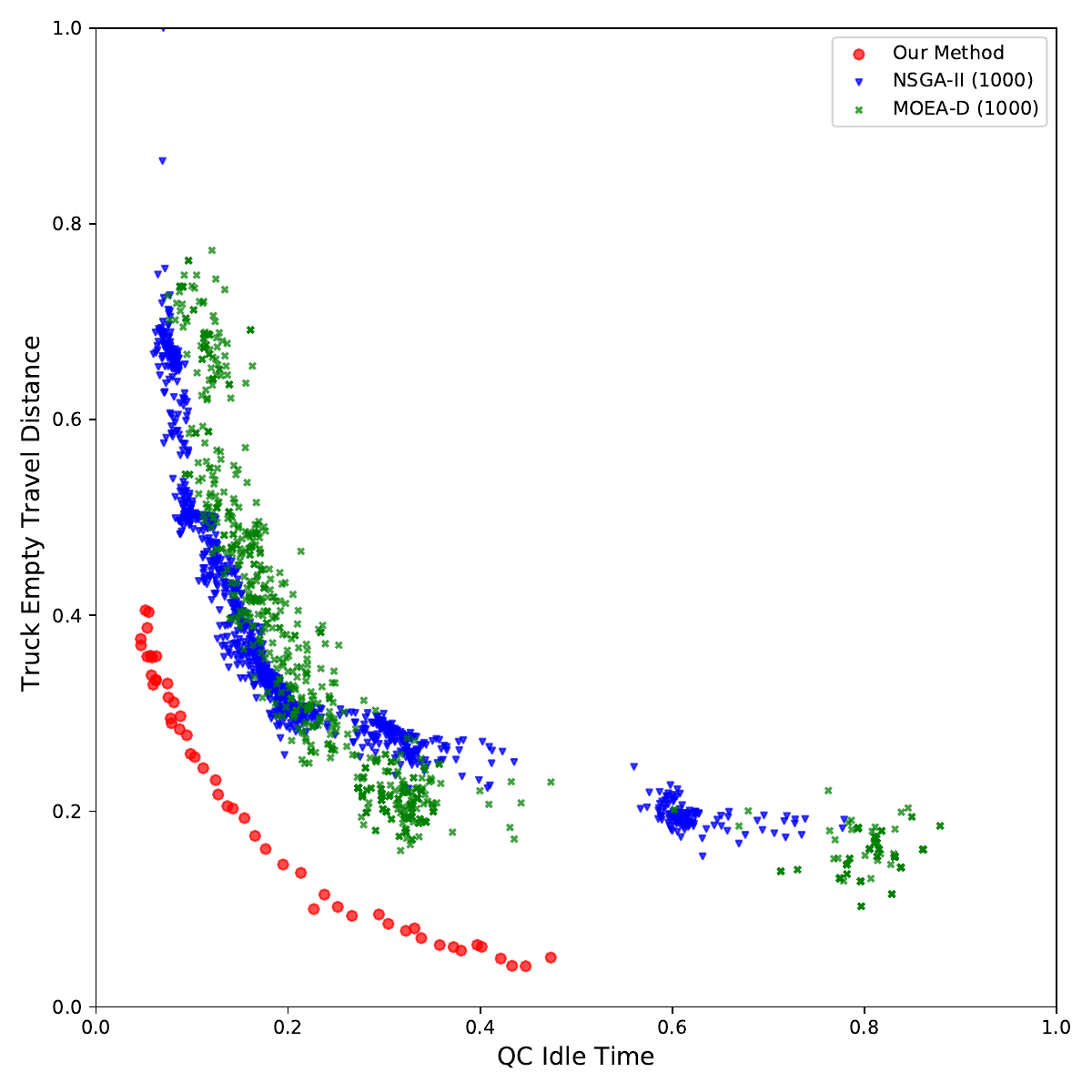} &
            \includegraphics[width=0.3 \linewidth]{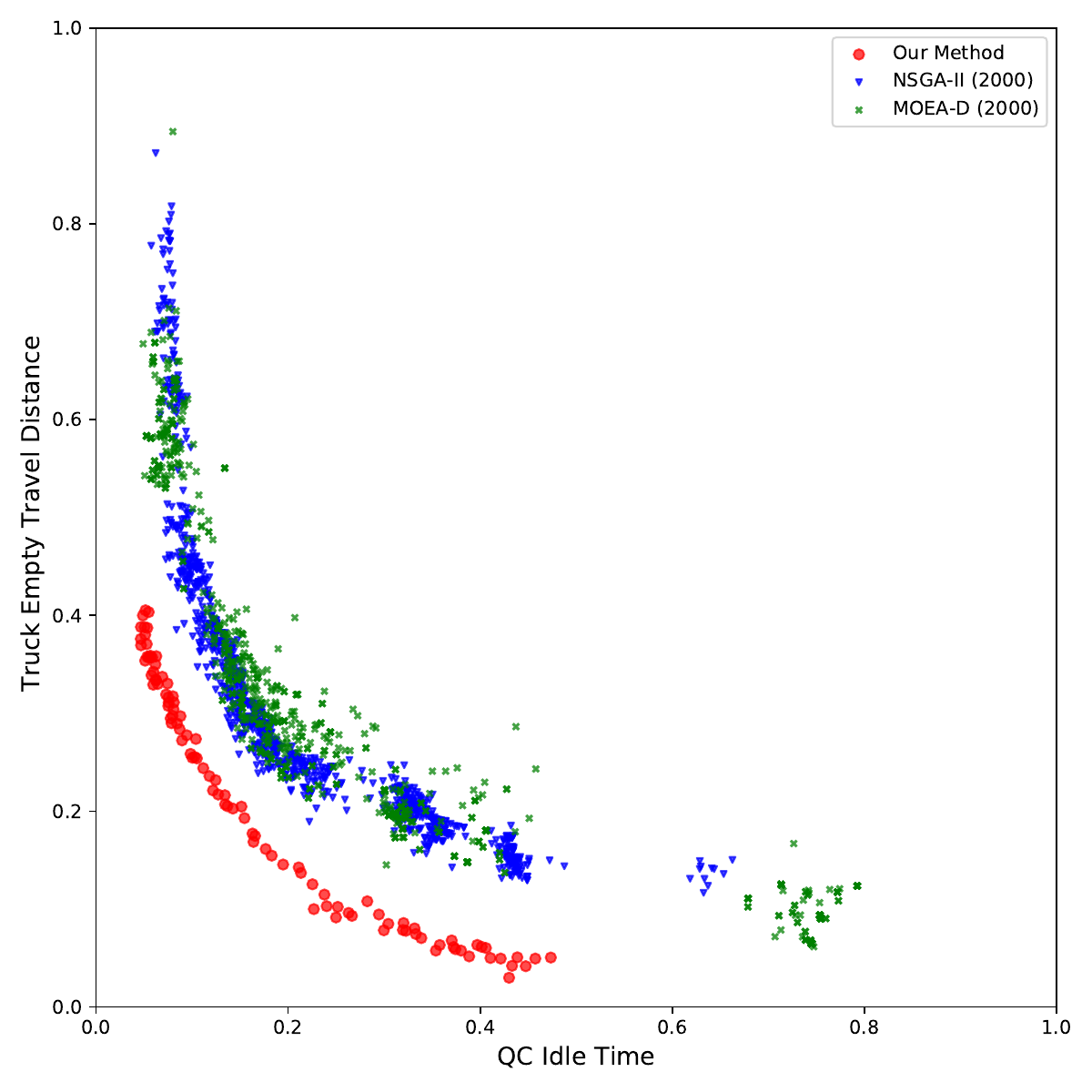} \\
            (d) 100 Trucks, 500 Generations & (e) 100 Trucks, 1000 Generations  & (f) 100 Trucks, 2000 Generations \\ \\
        \end{tabular}
    \end{minipage}
    \begin{minipage}[t]{1.0\linewidth}
    \centering
        \begin{tabular}{@{\extracolsep{\fill}}c@{}c@{}c@{}@{\extracolsep{\fill}}}
            \includegraphics[width=0.3 \linewidth]{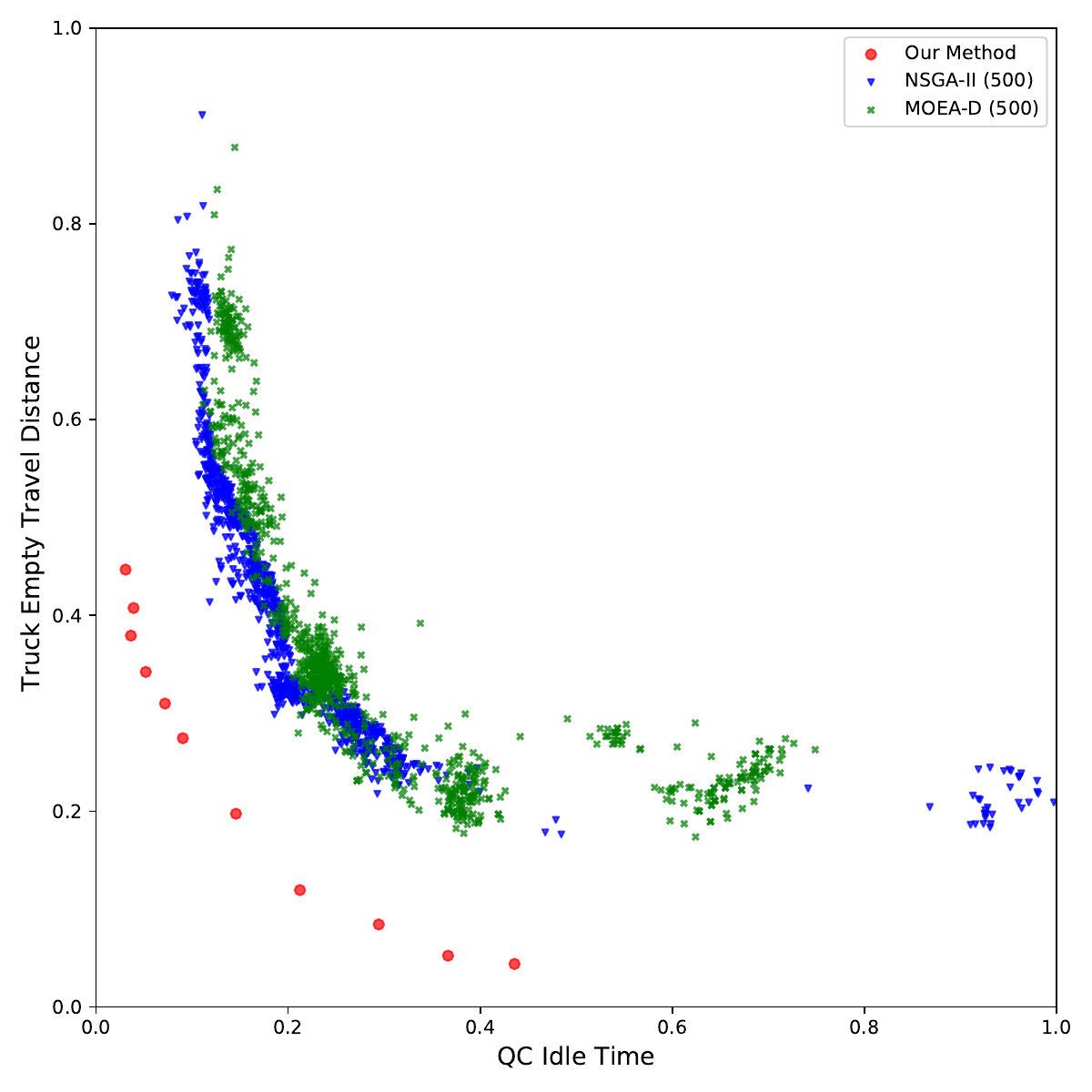} &
            \includegraphics[width=0.3 \linewidth]{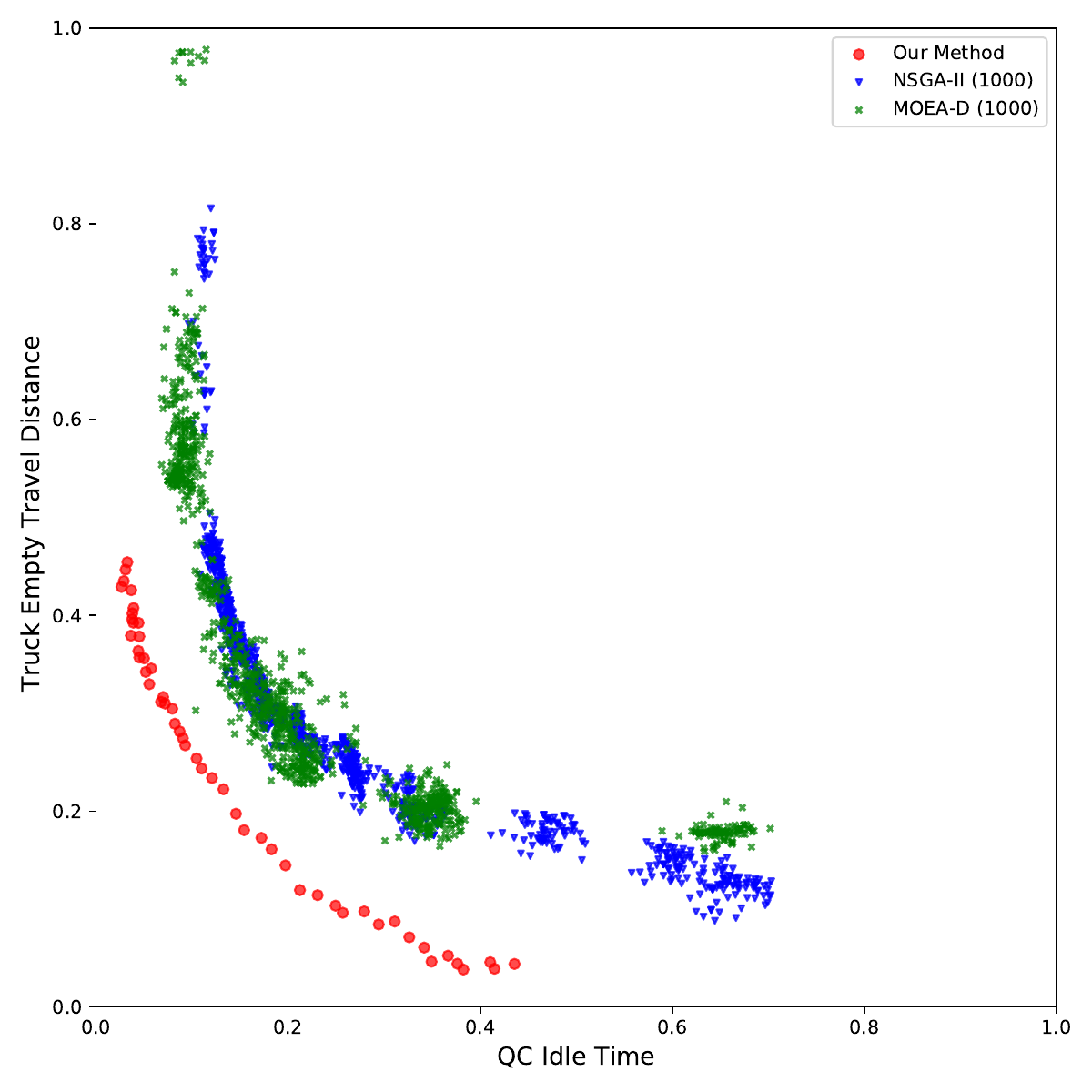} &
            \includegraphics[width=0.3 \linewidth]{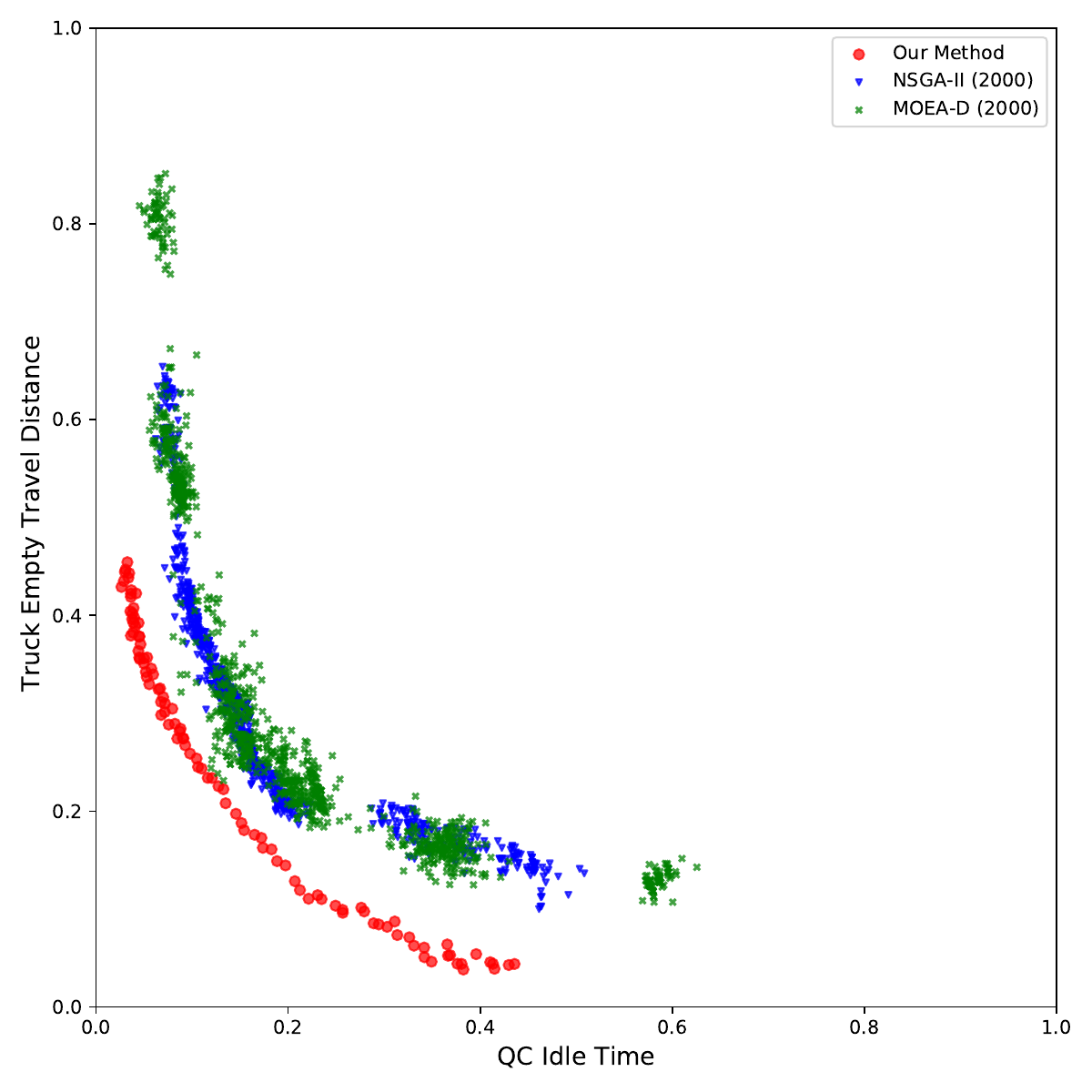} \\
            (g) 120 Trucks, 500 Generations & (h) 120 Trucks, 1000 Generations  & (i) 120 Trucks, 2000 Generations \\ \\
        \end{tabular}
    \end{minipage}
    \caption{Approximate Pareto front obtained by our method and benchmarks on instances of different number of trucks.}
    \label{fig:vis_pareto_fronts}
 \end{figure*}

\rev{To better understand the characteristics of our proposed method and factors that contribute its superior performance, we further visualize the obtained approximated Pareto frontiers in Figure~\ref{fig:vis_pareto_fronts}}. It can be seen that the approximate Pareto frontiers obtained by our method could completely cover those of the two evolutionary algorithms. In cases of 2000 generations, merely a few policies generated by evolutionary algorithms are close to or at the same level as our method (see Figure~\ref{fig:vis_pareto_fronts} (c) and (i)). Such outstanding results demonstrate the great potentials of RL-based methods for online MOO problems. \rev{We attribute this performance edge to the superior perceptual capacity of deep reinforcement learning, enabled by our customized neighborhood-aware attention module. By effectively capturing high-dimensional features and their intricate spatio-temporal relationships, DRL outperforms genetic-programming approaches that are restricted to a limited set of genetic operators and the size of decision trees. Therefore evolutionary algorithms exploit the raw feature data far less thoroughly.}


Apart from HV, the proposed PAMOO method also shows significant performance gains in terms of smoothness of the Pareto frontiers. The Pareto frontiers by the two evolutionary algorithms usually have large gaps (See Figures \ref{fig:vis_pareto_fronts} (d),(e) and (f)). A closer investigation reveals that a likely reason for this is the lack of effective fine-tuning mechanisms to accurately respond to small changes in preferences. Usually, a policy presented by a tree-based structure leads to large changes in its dispatching logic when modified by genetic operators (crossover, mutation). Such characteristic makes the policies tend to be the same for similar preferences, thus, forming the discontinuous intervals among the Pareto frontiers. In contrast, our method could still make the correct perception even though the preference is slightly changed because the preference embedding layer maps the preference to a longer feature vector (128 dimensions, same to the QC feature vector) which is adequate to discriminate similar preferences and influence the neural network. That is the reason why our method could generate much more continuous and even Pareto front compared to the benchmarks. In addition, the problem of "equivalent trees" (different trees have the same arithmetic logic) also exists for evolutionary algorithms and causes considerable policy overlap, which further worsens the discontinuity of the Pareto front.

\subsection{Performance of Generalization}

\begin{figure}[htbp]
\centering
\includegraphics[width=0.3 \textwidth]{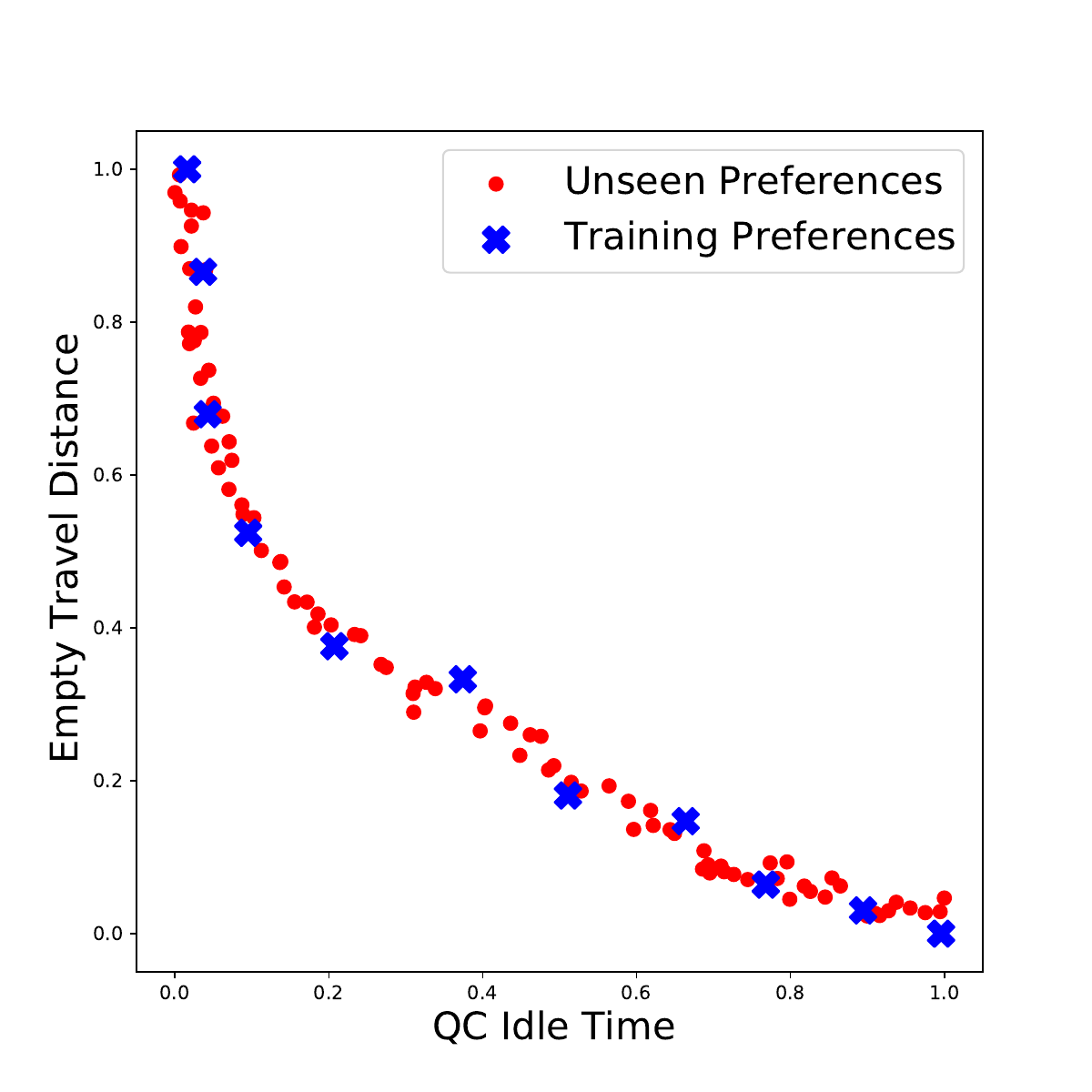}
\caption{\label{fig:prefs_gen} Pareto frontiers generated by proposed method that trained by a small number of preferences.}
\end{figure} 

We also conducted PAMOO's generalization experiment by using merely 11 guidance preferences on instances with 120 trucks. The results are evaluated against the PAMOO with 101 preferences. The PAMOO agent yields 40 non-dominated policies out of 101 input preferences.  Figure~\ref{fig:prefs_gen} visualizes the approximated Pareto front. According to the result, the generalized policies could fill up the intervals among the policies with training preferences over the objective space and all policies approximate a relatively dense Pareto front. Like most of the machine learning approaches, fewer training samples (preferences in our case) lead to certain degrees of performance drops. \rev{PAMOO} is able to generate 11 non-dominant policies out of 11 input preferences if the evaluating preferences are same to the training instances but when 90 unseen evaluating instances are added, PAMOO could only generate 29 more non-dominated policies. This problem could be alleviated by taking more preferences into the training process and increasing the network size. In general, the generalization performance is crucial for the proposed methodology as it makes it possible for the agent to dispatch under arbitrary preferences and allow users interactively adjust the preferences in real-world cases.

\subsection{Comparison with Outer Loop Methods}\label{sec:outer_loop}

\begin{figure}[htbp]
\centering
\includegraphics[width=0.3 \textwidth]{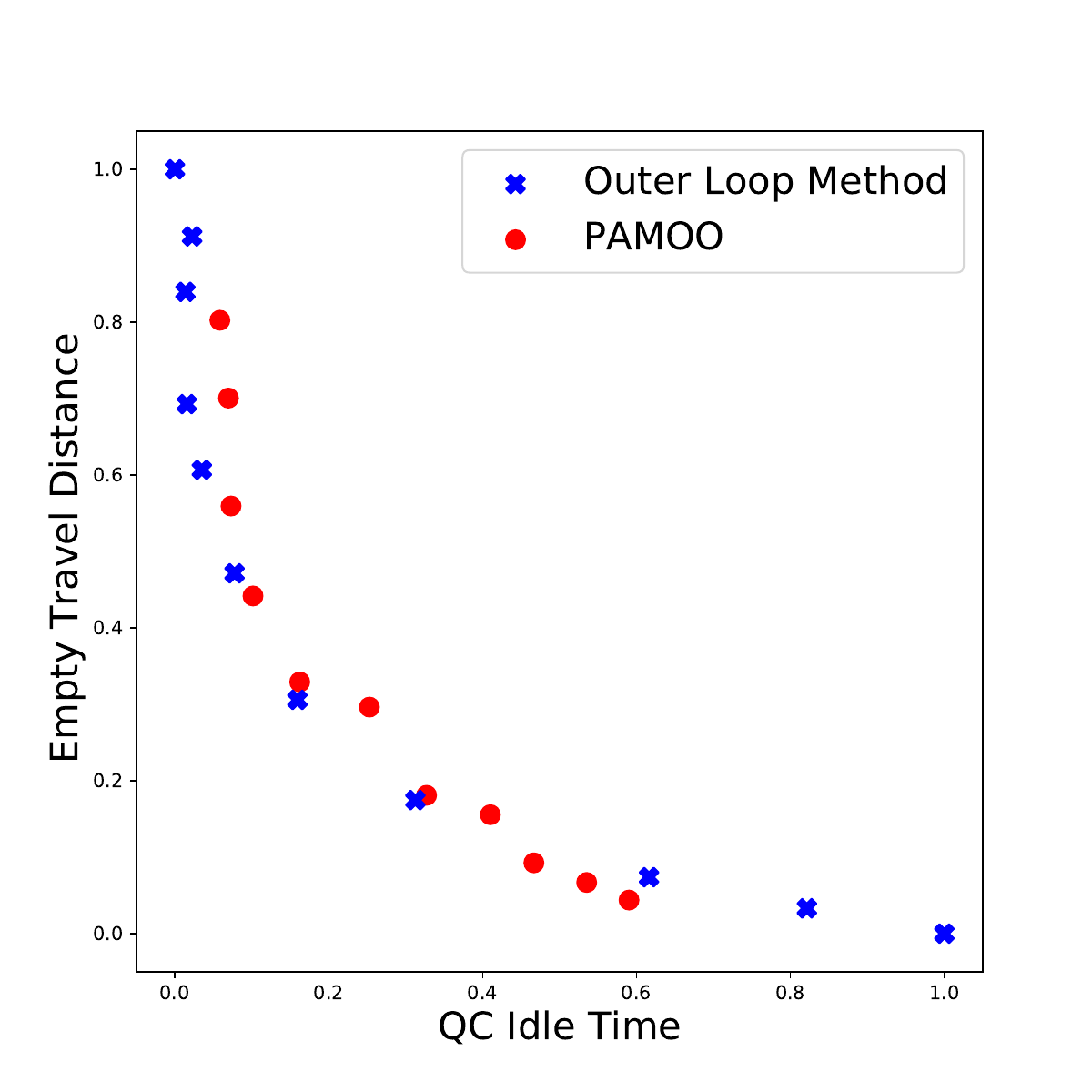}
\caption{\label{fig:inner_outer_loop} Pareto frontiers generated by PAMOO and outer loop method on instance of 120 trucks.}
\end{figure}

We compare the performance of PAMOO with an outer loop method. As introduced in the literature, the outer loop methods consider a multi-objective optimization problem as several single-objective optimization problems with different preferences. In other words, an outer loop method needs to train separate dispatching policies for different preferences. In this experiment, the same network structure except the eliminated preference embedding layer is used for training single-objective policies. Since training a single policy from scratch is time-consuming, only 11 policies are trained separately for comparison in this experiment. Figure~\ref{fig:inner_outer_loop} shows the approximated Pareto frontiers on 11 evenly distributed preference weights. It can be seen that the outer loop method generates comparable Pareto frontiers measured to PAMOO in terms of HV (0.786 and 0.784, respectively) and the gap is tiny. Bear in mind that our proposed PAMOO method only uses one single network as MOO policies. However, the outer loop method has to be trained separately at different preferences which is extremely time costly.  

Visually, the approximated Pareto front obtained by the outer loop method covers broader space. This property is further analyzed by numerical results of each preference which are presented in Table \ref{tab:inner_outer_result}. In extreme cases (preferences of [1.0, 0.0] or [0.0, 1.0]), the examined problem degenerates into a single-objective problem and the outer loop method shows better performance. The outer loop method achieved the range of 15.08 min/QC and 393.86 m/task for both objectives while our method obtained the ranges of 8.02 min/QC and 298.77 m/task respectively, having gaps of 46.8\% and 24.1\%. In general, at extreme cases, one of the objectives is further optimized to a little degree at the expense of dramatic performance drop in the other objective.

\begin{table*}[htbp]
\centering
\caption{\label{tab:inner_outer_result}Numerical Result of Comparison to Outer Loop Method.}
\begin{tabular}{|c | c c | c c |}
\hline
\multirowcell{3}{Preferences} & \multicolumn{2}{c|}{PAMOO} & \multicolumn{2}{c|}{Outer Loop Method} \\
\cline{2-5}
  & \multirowcell{2}{QC Idle Time \\ (min/QC)} & \multirowcell{2}{Empty Travel Distance \\ (m/task)} & \multirowcell{2}{QC Idle Time \\ (min/QC)} & \multirowcell{2}{Empty Travel Distance \\ (m/task)}  \\ &  &  &  & \\
\hline
[1.0, 0.0] & 13.04 & 799.11 & 12.16 & 876.9 \\

[0.9, 0.1] & 13.21 & 759.0 & 12.49 & 842.33 \\

[0.8, 0.2] & 13.26 & 703.38 & 12.36 & 813.71 \\

[0.7, 0.3] & 13.69 & 657.03 & 12.39 & 755.92 \\

[0.6, 0.4] & 14.6 & 612.71 & 12.68 & 722.04 \\

[0.5, 0.5] & 15.97 & 599.77 & 13.32 & 668.71 \\

[0.4, 0.6] & 17.09 & 554.33 & 14.56 & 603.52 \\

[0.3, 0.7] & 18.34 & 544.31 & 16.87 & 551.84 \\

[0.2, 0.8] & 19.19 & 519.52 & 21.45 & 512.14 \\

[0.1, 0.9] & 20.23 & 509.42 & 24.55 & 496.1 \\

[0.0, 1.0] & 21.06 & 500.34 & 27.24 & 483.04 \\

\hline
\end{tabular}
\end{table*}

\begin{figure}[htbp]
\centering
\includegraphics[width=0.3 \textwidth]{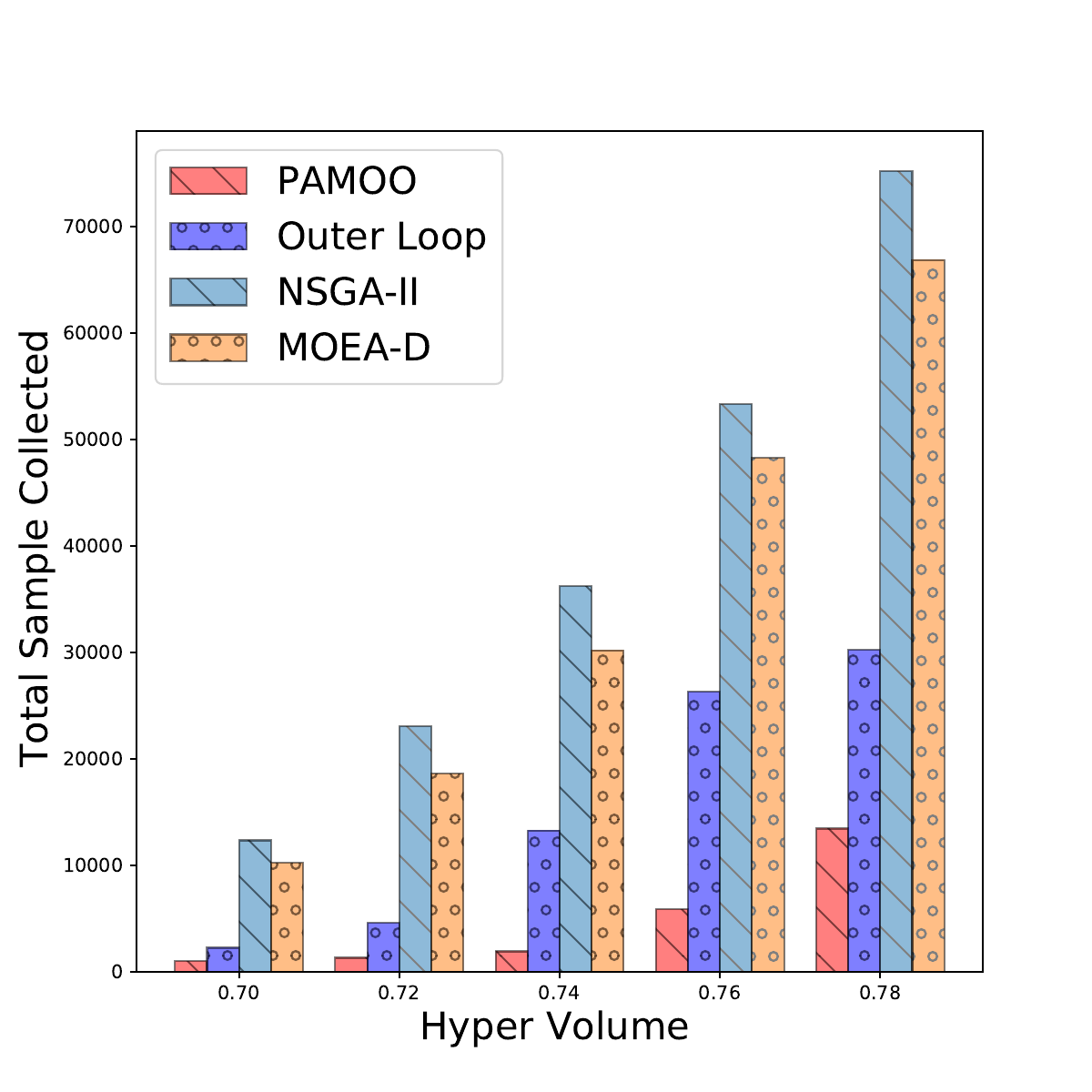}
\caption{\label{fig:sample_efficiency}  Sample efficiency of inner-loop (ours) and outer-loop methods compared with NSGA-II and MOEA-D.}
\end{figure}

Another advantage of the proposed methodology is the higher sample efficiency than the outer loop method and evolutionary algorithms. A sample indicates one single run for a particular problem instance and is equivalent to an episode in RL training process. The Figure~\ref{fig:sample_efficiency} shows the total number of samples (episodes) that are required for these algorithms as HV increases. Generally speaking, the evolutionary algorithms (NSGA-II and MOEA/D) need to collect more than 5 times of samples than the proposed PAMOO method. For both evolutionary algorithms, their population sizes are large and numerous samples are required to evaluate the fitness of each individual. For RL-based approaches, our proposed method requires less than half of the episodes compared to the outer loop method. This is probably because the features extracted by the network at different preferences are correlated and hence learning is transferred across different preferences. For example, the dispatching policy under preference of [0.5, 0.5] must be somehow similar to the policies under preferences of [0.4, 0.6] or [0.6, 0.4]. Therefore, when the network is trained for preference [0.5, 0.5], it is also partially trained for the preferences [0.4, 0.6] and [0.6, 0.4]. More precisely, it is believed that when the network is trained for a preference, it is actually trained for all other possible preferences simultaneously to some extent. This also explains why our proposed method can produce solutions at preferences that are not seen in the training. 

\subsection{Ablation Study}
\rev{To validate the efficacy of our key architectural and algorithmic design choices, and to gain a deeper understanding of how each component contributes to multi-objective optimization performance of the examined problem, an ablation study is conducted to investigate the impact of both the target preference encoding strategy and the reinforcement learning algorithm.} 

\begin{figure}[htbp]
\centering
\includegraphics[width=0.3 \textwidth]{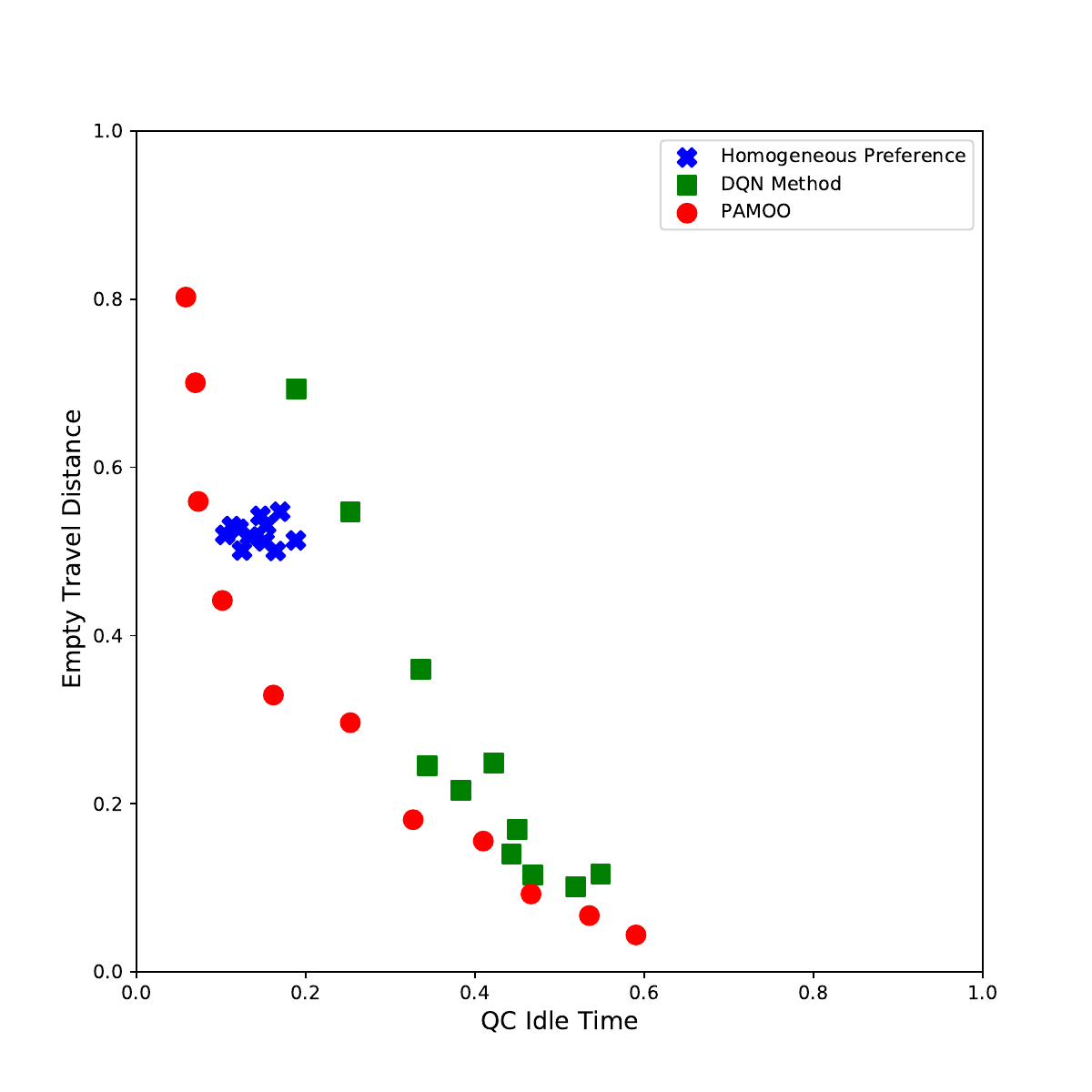}
\caption{\label{fig:ablation_study_result} Ablation study results on adopting various RL algorithms and network architectures.}
\end{figure}

\rev{Figure~\ref{fig:ablation_study_result} visualizes the results of this ablation study. Our proposed PAMOO approach (red dots), delivers the most robust and comprehensive Pareto frontier, exhibiting a broad and well-distributed range of trade-offs that reflect its ability to effectively balance both optimization objectives in this experiment. By contrast, the DQN-based variant (green squares) demonstrates a clear bias: while it achieves relatively stronger performance on the vehicle empty travel distance objective, its capacity to optimize QC idle time is notably inferior to PAMOO. This results in a narrower, less optimal Pareto frontier, highlighting a substantial performance gap that underscores the superiority of PPO over DQN for this dynamic, multi-objective problem setting.}

\rev{For the ablation variant (blue crosses), where target preferences are treated as generic observation features rather than being embedded separately and fused via Hadamard product, solutions cluster tightly in a restricted region of the frontier, focusing almost exclusively on minimizing QC idle time. Even though this variant outperforms the DQN variant on this single objective, its failure to explore meaningful trade-offs across both objectives confirms that the target preference input failed to exert its intended regulatory effect, leading the model to degenerate into a single-objective optimization process. These findings not only validate the critical role of our dedicated preference embedding and fusion mechanism in enabling robust multi-objective exploration but also provide empirical evidence that our design mitigates the risk of objective degeneracy, a key concern in real-world multi-objective scheduling systems.
}

\rev{PAMOO’s performance of full Pareto coverage hinges on the Hadamard product in the network design, which acts as a dimension-wise scaling factor. It preserves intrinsic crane state information while precisely modulating objective-relevant feature intensity via preferences, while the preference feature fail to work as a homogeneous component of state observation since its significance is diluted during backpropagation.}

\subsection{Preference Calibrations for PAMOO}\label{sec:preference-calib}
\rev{The calibration function is applied to preference weights before the weights are fed to the preference embedding module.} 
The experiment of preference calibration is conducted on instances with 80, 100 and 120 trucks. The specific adjusting method based on a set of existing Pareto policies is described in section \ref{algorithm}. Firstly, a set of Pareto policies is generated by a list of evenly distributed preferences. Then, each preference is adjusted by using the aforementioned preference calibration method based on the relative locations of these policies in the objective space. \rev{This process iterates until we have obtained sufficient number of policies that have strong alignment with corresponding preference directions. We then take them as \texttt{anchor policies} for calibrating any future dynamic user input preference vector. This would ensure minimum computation during PAMOO inference and hence enable real-time dispatching.}

Table \ref{tab:adjust_prefs} shows the PAMOO results with and without calibration. In our experiments, the number of preferences is set to different values (11, 21 and 51) to evaluate the trends of the performance gains through sampling. It can be seen that, at the same number of preferences, PAMOO with calibration achieves better HV scores than PAMOO without calibration. It is axiomatic that more anchor preferences contribute to higher HV as well. When the number of anchor preferences reaches 51, PAMOO with calibration obtain the best results among all. 
The improvements range between 0.4\% and 1.79\% with smaller improvements being obtained when the number of preferences is relatively large since the optimization space for adjusting preferences is quite limited. 
%
%
The uneven and irregular Pareto front is probably caused by uneven sensitivities in objectives for the same degree of preference change. 
 According to Table \ref{tab:inner_outer_result}. There is a moderate change of 0.65 for PAMOO with the preference changing from [1.0, 0.0] to [0.7, 0.3] but when preference weights are changed from [0.3, 0.7] to [0.0, 1.0], the change in the QC idle time is 2.72, which is much larger. For outer loop method, the difference is even larger (0.23 and 10.37, respectively). which is not ideal. There could be multiple contributing factors to this phenomenon. For example, the number of trucks, task distribution among QCs, and the relative locations of import-export yards, all of these factors could make optimizing one objective harder than the other, thus causing the final approximated Pareto front uneven. It is worth conducting a systematic sensitivity analysis of these factors and their interconnections in future.

\begin{table*}[htbp]
\centering
\caption{\label{tab:adjust_prefs}Comparison Result between Even and Adjusted Preferences.}
\begin{tabular}{|c | c c c | c c c | c c c|}
\hline
\multirowcell{2}{Method} & \multicolumn{3}{c|}{80 Trucks} & \multicolumn{3}{c|}{100 Trucks} & \multicolumn{3}{c|}{120 Trucks} \\
  & HV & Sparsity & Gap(\%) & HV & Sparsity & Gap(\%) & HV & Sparsity & Gap(\%) \\
\hline
Evenly Preferences (11 Pref.) & 0.684 & 253.03 & 5.79\% & 0.676 & 260.71 & 7.02\% & 0.679 & 267.98 & 7.24\% \\

Evenly Preferences (21 Pref.) & 0.709 & 64.48 & 2.34\% & 0.705 & 66.14 & 3.03\% & 0.709 & 68.91 & 3.14\% \\

Evenly Preferences (51 Pref.) & 0.723 & 11.42 & 0.41\% & 0.721 & 11.8 & 0.83\% & 0.724 & 11.06 & 1.09\% \\
\hline
Adjusted Preferences (11 Pref.) & 0.694 & 240.93 & 4.41\% & 0.683 & 240.13 & 6.05\% & 0.685 & 241.41 & 6.42\% \\

Adjusted Preferences (21 Pref.) & 0.716 & 58.29 & 1.38\% & 0.718 & 60.55 & 1.24\% & 0.719 & 60.9 & 1.78\% \\

Adjusted Preferences (51 Pref.) & 0.726 & 10.28 & 0.0\% & 0.727 & 10.29 & 0.0\% & 0.732 & 10.07 & 0.0\% \\
\hline
\end{tabular}
\end{table*}


\section{Conclusions}\label{sec:conclusion}
  
In this research, we proposed the first learning based dynamic multi-objective approach for real-life vehicle dispatching in marine container terminals. Benefiting from the innovative network structure design and preference calibration, the method has short response time, superior ability to overcome uncertainties in the environment and also accurate and diverse trade-off policies for users.
The experiments against traditional evolutionary multi-objective algorithms based on genetic programming demonstrated that our proposed method could outperform the benchmarks on solution quality, diversity and sample efficiency.    


\rev{The proposed methodology could be extended in several ways. First, future research could explore more accurate preference embedding mechanisms for MOO objectives with varying degrees of sensitivity to preference changes, which may produce uneven and irregular Pareto fronts. Second, the experimental settings in this paper are limited. Future studies could investigate complex QC–Yard matching scenarios to fully assess the capabilities of the PAMOO algorithm. Finally, it is worth investigating alternative feature fusion mechanisms and network architectures (such as transformers and MoEs) to eliminate the need for post-training preference calibration and to further improve the performance.}

\bibliographystyle{cas-model2-names}
\bibliography{references}

\end{document}